\documentclass{style}

\usepackage{amsmath,amssymb,mathtools}
\usepackage{algorithm}
\usepackage{algpseudocode}
\DeclareRobustCommand{\recongptmark}{\includegraphics[height=1.55ex]{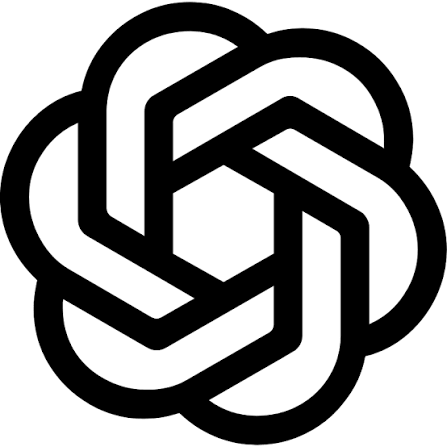}}
\usepackage{tabularx}
\usepackage{fontawesome5}
\usepackage{inconsolata}
\usepackage{xspace}
\usepackage{xurl}

\newcommand{\method}{\textbf{\textsc{RoboExpand}}\xspace}
\newcommand{\papertitle}{\textsc{RoboExpand}: A Real2Sim2Real Data Engine with Agentic Reconstruction and Multi-View Video Generation}

\hypersetup{pdftitle={RoboExpand: A Real2Sim2Real Data Engine with Agentic Reconstruction and Multi-View Video Generation}}
\renewcommand{\titlefont}{\fontsize{16}{20}\selectfont}
\title{\rmfamily\papertitle}

\renewcommand{\authorlist}{%
  \authorfont\rmfamily\mdseries
  Yifan~Luo$^{1,3,*}$, Xin~Wu$^{2,3,*}$, Xiancong~Ren$^{3,\dag}$, Chenhui~Deng$^{3,4}$, Huimin~Ma$^{3}$, Jianxin~Sun$^{3}$, \\
  Yifan~Sun$^{3,5}$, Qinfan~Zhang$^{3,4}$, Han~Dong$^{3}$, Jin~Xu$^{3}$, Zhiheng~Liu$^{3}$, \\
  Chen~Sun$^{2}$, Yong~Dai$^{3}$, Bing~Su$^{1,\ddag}$, Xiaozhu~Ju$^{3,\ddag}$%
}

\renewcommand{\affiliationfont}{\fontsize{8.4}{10}\selectfont}
\renewcommand{\affiliationlist}{%
  \affiliationformat[1]{Renmin University of China}\,\textperiodcentered\,
  \affiliationformat[2]{The University of Hong Kong}\,\textperiodcentered\,
  \affiliationformat[3]{Beijing Innovation Center of Humanoid Robotics}\\[1pt]
  \affiliationformat[4]{Beihang University}\,\textperiodcentered\,
  \affiliationformat[5]{China University of Mining \& Technology, Beijing}%
}

\contribution[*]{Equal Contribution}
\contribution[\dag]{Project Leader}
\contribution[\ddag]{Corresponding Author}

\abstract{%
Collecting real-robot demonstrations across varied layouts and visual conditions is costly. Simulation can broaden this coverage, but discrepancies in geometry, contact dynamics, and appearance often hinder sim-to-real policy transfer.
We introduce \textbf{\method}, an agentic Real2Sim2Real data engine that expands a single real-robot demonstration into diverse action trajectories and visual observations for policy learning.
A reconstruction agent first replays the recorded actions and uses discrepancies in contact and object motion to refine the scene's geometry and physics. A trajectory generalization agent then adapts the demonstrated motion to new object layouts by planning paths that account for the new object positions. For each successful trajectory, the engine's learned multi-view transfer model finally generates synchronized videos under different visual conditions, preserving alignment with the trajectory's actions and states.
Across four tasks on three real robot platforms, with 80 demonstrations per task, policies trained on \method-generated data achieved an average success rate of 72.5\%, up from 40.0\% with raw simulator renderings and close to the 75.0\% achieved with real demonstrations.
In additional generalization experiments on two Astribot tasks, policies trained on synthetic data covering the five tested visual conditions achieved an average success rate of 68.0\%, compared with 39.0\% for policies trained on real demonstrations from the default scene.
}

\checkdata[\faDesktop~Website]{\url{https://roboexpand.github.io}}
\checkdata[\faGithub~Code]{\url{https://github.com/RoboExpand/RoboExpand}}
\fancypagestyle{reportfirst}[firststyle]{%
  \fancyhead[L]{\vskip 3mm
    \raisebox{0.75mm}[11.5mm][0pt]{%
      \includegraphics[trim=346bp 34bp 346bp 34bp,clip,height=10mm]{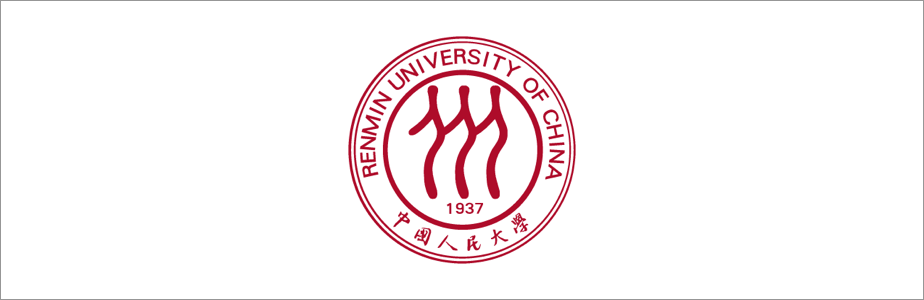}%
      \hspace{5mm}%
      \includegraphics[trim=195bp 45bp 197bp 46bp,clip,height=10mm]{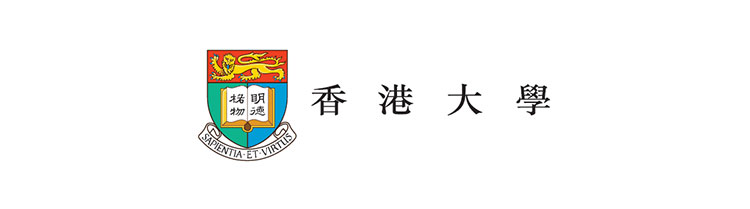}%
    }%
  }%
  \fancyhead[R]{\vskip 3mm\includegraphics[height=11.5mm]{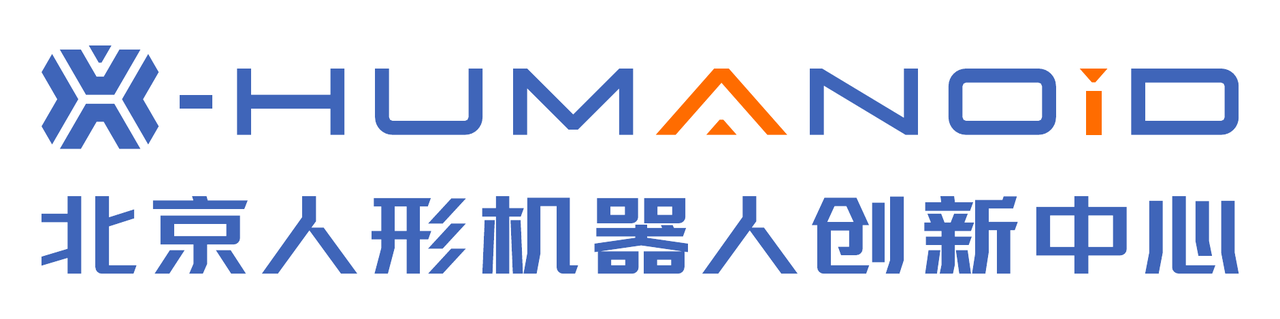}}%
}

\definecolor{frontlinkblue}{HTML}{3977AC}

\begin{document}
\hypersetup{urlcolor=frontlinkblue}
\maketitle
\hypersetup{urlcolor=spatheme}
\raggedbottom
\thispagestyle{reportfirst}

\section{Introduction}
\label{sec:intro}

The scarcity of realistic and diverse interaction data remains a major barrier to generalization in embodied AI~\citep{openx2024,mandlekar2022robomimic}. Each real-robot demonstration provides a valuable record of observations, actions, and their physical consequences, yet captures only one execution of a task under a limited set of conditions. Broadening this coverage across objects, scenes, and task settings requires repeated real-world collection, demanding substantial human effort, robot time, and hardware resources~\citep{khazatsky2024droid}. These costs make it difficult to acquire the breadth of experience needed for robust generalization. This creates a pressing need for scalable approaches that expand limited real-world experience into diverse training data while preserving physically plausible interactions and reliable alignment between observations and actions.

\begin{figure}[!htbp]
  \centering
  \includegraphics[width=0.99\linewidth]{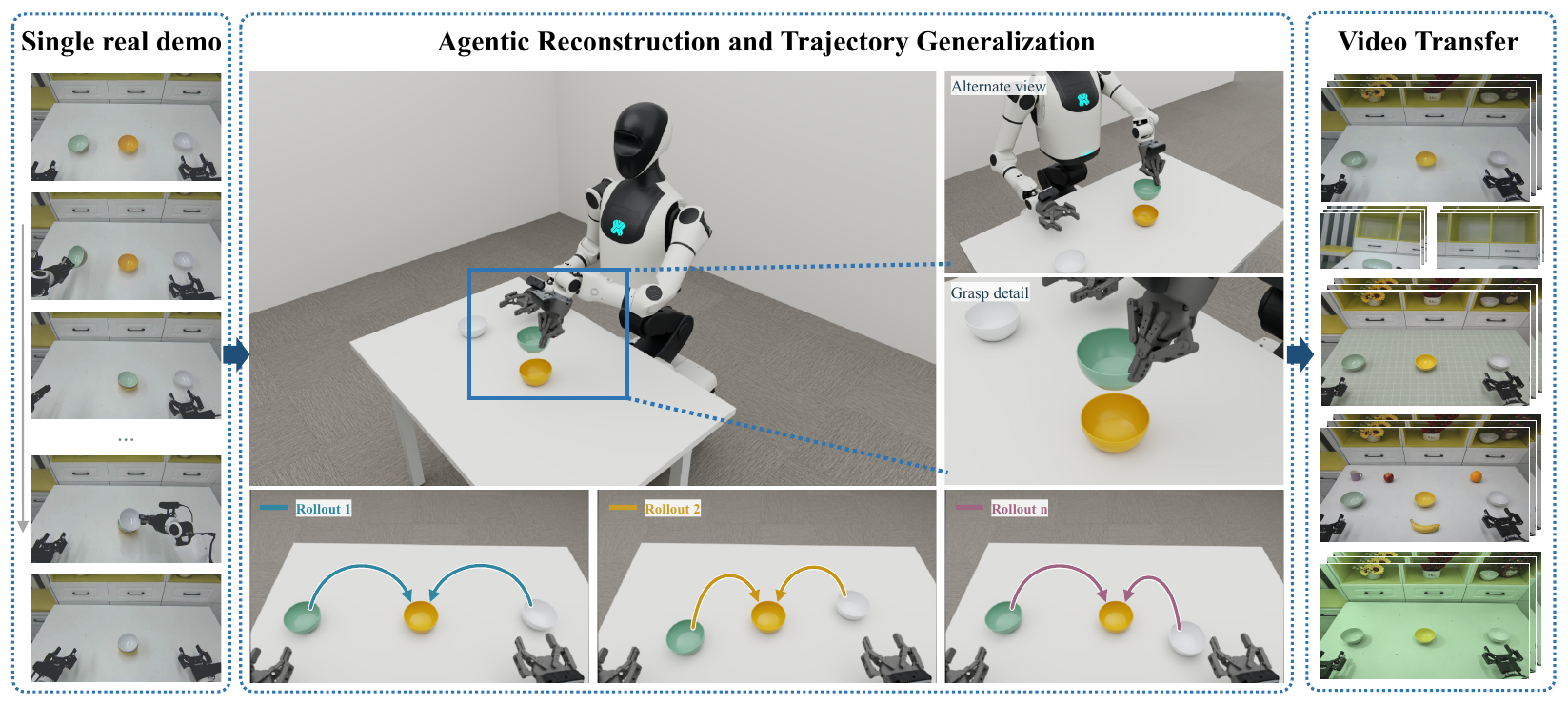}
  \caption{\textbf{From a single real demonstration to diverse robot training data.} Starting from one real demonstration \textbf{(left)}, \method reconstructs an interactive simulation scene and generates multiple trajectories under varied object layouts \textbf{(center)}. Each trajectory is then transferred into realistic multi-view videos with diverse visual conditions \textbf{(right)}, yielding both motion and appearance diversity while retaining aligned robot actions and states.}
  \label{fig:teaser}
\end{figure}

Simulation offers a promising route to expanding interaction data through repeatable execution in configurable environments~\citep{mandlekar2023mimicgen,chen2026robotwin2}. Real2Sim2Real grounds this process in real scenes and demonstrations, using simulation to expand training experience for subsequent real-world policy learning~\citep{torne2024rialto,fang2025rebot}. Turning this potential into effective training data requires meeting two complementary requirements. First, simulated interactions must be faithful to the observed behavior and adaptable to new conditions. Image-based reconstruction can recover scene geometry and articulation~\citep{chen2024urdformer}, but visual agreement alone does not establish whether recorded actions reproduce the observed object responses or whether adapted trajectories remain executable under new object layouts. Second, visual diversification must preserve the underlying interaction~\citep{vla-trace}. Generative Sim2Real transfer methods provide increasingly realistic and diverse observations~\citep{yu2025lucidsim,huang2026roborender}, whose value for policy learning depends on their alignment with simulated motion and action supervision across time and cameras~\citep{zhang2026geomemvlainternalizing3dscene}. These requirements motivate a data generation pipeline that connects interaction-grounded reconstruction and trajectory adaptation with consistent visual diversification.

To unify interaction generation and visual diversification, we introduce \method, an agentic Real2Sim2Real data engine that expands a single real-robot demonstration into diverse training data for real-world policy learning (\cref{fig:teaser}). A reconstruction agent builds an interactive scene and refines its task-relevant geometry and physical properties by replaying the recorded actions and comparing simulated object motion with that observed in the demonstration. Within the reconstructed scene, a trajectory generalization agent expands interaction coverage by adapting the demonstrated motion to new object layouts, reusing feasible segments and replanning those affected by changes in object placement. For each successful simulated trajectory, a multi-view video transfer model uses simulated depth to generate synchronized camera views under varied visual conditions. Each visual variant is paired with the actions and states recorded at the corresponding simulation steps, producing visually diverse training sequences with supervision from the simulator.

We evaluate downstream policy learning on four tasks across Astribot, AgileX, and Tienyi 2.0, using 80 training demonstrations per task for each data source. Policies trained on \method-generated data achieve a mean real-world success rate of 72.5\%, compared with 40.0\% for policies trained on reconstructed simulation data and 75.0\% for those trained on real-robot data. On two Astribot tasks evaluated under five visual variations, policies trained on synthetic data covering these variations achieve 68.0\% mean success, compared with 39.0\% for policies trained on default-scene real data. In video generation evaluations, our model achieves the best video reconstruction scores and mean human-preference rank among the four compared methods.

Our contributions are summarized as follows:
\begin{itemize}
    \item We recast a single real-robot demonstration as both a reference for reconstructing interactive scenes and a seed for generating diverse training data. Our agentic Real2Sim2Real data engine, \method, connects these capabilities with downstream policy learning and real-world evaluation in a unified workflow.

    \item We combine action-replay feedback for scene refinement with segment reuse and selective replanning for trajectory adaptation. The resulting trajectories supply both depth conditions for synchronized multi-view generation and corresponding action supervision, keeping visual diversification grounded in executable interactions.

    \item We demonstrate that the generated data supports effective policies across tasks and robot platforms, with performance approaching real-data training. These findings support broader reuse of collected demonstrations to expand training coverage, offering a practical route toward reducing repeated task-specific real-robot collection.
\end{itemize}

\section{Related Work}
\label{sec:related_work}

\noindent\textbf{Synthetic Data for Embodied Learning.}
Synthetic data expands observations, behaviors, and environments for embodied learning, reducing reliance on real-robot data collection~\citep{mandlekar2023mimicgen,xue2025demogen}.
Simulation-based pipelines execute adapted demonstrations or generated task programs~\citep{mandlekar2023mimicgen,chen2026robotwin2}, while digital-cousin construction and asset reconstruction expand their environment coverage~\citep{dai2025acdc,zhang2026autorsr}.
Physics-based rollouts provide action/state supervision but depend on simulation-ready assets~\citep{xiang2020sapien,mandlekar2023mimicgen}. RialTo refines policies in reconstructed digital twins~\citep{torne2024rialto}, while ReBot combines replayed simulation motion with inpainted real backgrounds~\citep{fang2025rebot}.
Alternatively, geometry-based synthesis jointly transforms recorded point-cloud observations and actions~\citep{xue2025demogen} or renders reconstructed objects along demonstration-derived motion~\citep{yu2025real2render2real}, avoiding object dynamics simulation while relying on captured geometry and example behavior.
Generative visual synthesis can broaden appearance coverage while retaining simulator-derived supervision~\citep{yu2025lucidsim}; video-first synthesis can expand behavior coverage using inferred action labels~\citep{jang2025dreamgen}.
Dream4ACT~\citep{zhu2026dream4actsharedvisualaction} jointly models observations and actions across embodiments using rendered action views.
Our pipeline couples agentic asset reconstruction, seed-guided trajectory expansion, and multi-view visual transfer, generating diverse observations paired with simulated actions and states.

\noindent\textbf{Sim-Ready 3D Reconstruction.}
Simulation-ready asset creation requires geometric fidelity together with representations of how objects move and interact~\citep{xiang2020sapien,chen2024urdformer}.
Text- and image-conditioned 3D generation produces geometry and appearance~\citep{xiang2025trellis}; simulation-oriented prediction adds kinematic hierarchies~\citep{chen2024urdformer} and physical attributes~\citep{cao2026physxanything} in executable object or scene descriptions.
Complementing these representations, agentic pipelines refine candidates through feedback: visual critics guide articulated modeling~\citep{le2025articulateanything}, scene-level checks assess layouts and physical stability~\citep{xia2026sage}, and simulated interactions reveal behavioral errors and guide revisions to geometry or material models~\citep{xie2026codimrecon}.
Our reconstruction agent uses replayed robot actions to revise task-relevant geometry and contact behavior before trajectory expansion.

\noindent\textbf{Video Generation for Sim2Real Transfer.}
Controllable video generation seeks to vary appearance while preserving motion and geometry for robot learning~\citep{nvidia2025cosmostransfer,liu2025robotransfer}. Spatially conditioned image diffusion~\citep{zhang2023controlnet} and joint multi-view image diffusion~\citep{shi2024mvdream} provide related foundations for structural control and view consistency.
Masked video editing uses observed robot--object regions as anchors while varying their surroundings~\citep{wang2026robovip}, whereas geometry-conditioned rendering synthesizes observations from depth, normals, or semantic maps~\citep{nvidia2025cosmostransfer,liu2025robotransfer}.
The former retains visual evidence of interactions; the latter permits scene-wide appearance changes through structural guidance.
For synchronized cameras, spatial or latent concatenation shares information across views~\citep{liu2025robotransfer,huang2026roborender}, while occupancy-guided view attention models cross-camera dependencies~\citep{yang2025orv}.
History-conditioned continuation extends generation beyond fixed video windows~\citep{huang2026roborender}.
Our generator inserts cross-view residual modules into a shared depth- and reference-conditioned backbone, producing synchronized observations aligned with simulator action/state timestamps.

\begin{figure}[!htbp]
  \centering
  \includegraphics[width=\textwidth]{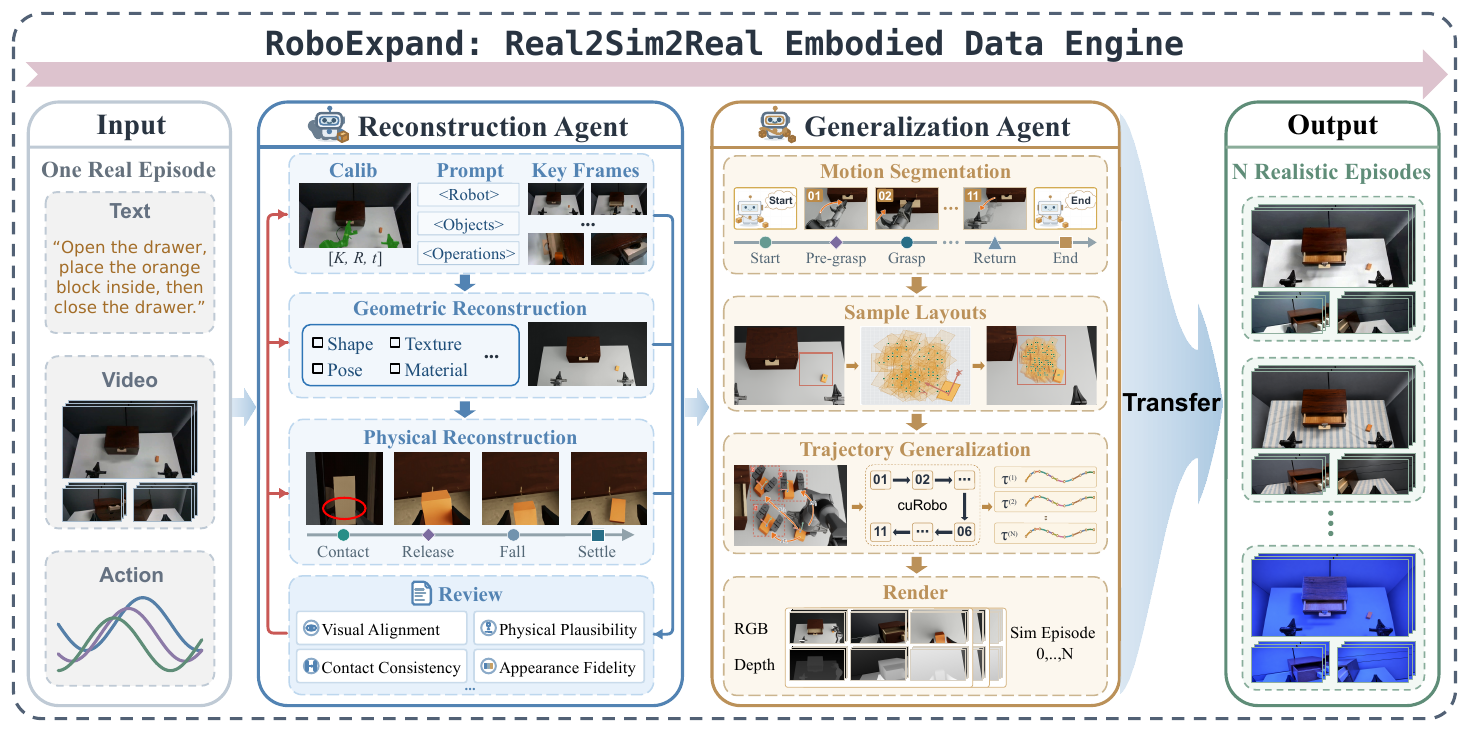}
  \caption{\textbf{Overview of \method.} Given a single real-robot episode, the \textbf{reconstruction agent} builds and refines an interactive simulation scene through geometric and physical reconstruction. The \textbf{generalization agent} adapts the demonstrated motion to new object layouts by reusing feasible motion segments and replanning others. \textbf{Multi-view video transfer} converts successful simulated rollouts into realistic visual variants, paired with aligned actions and states for policy training.}
  \label{fig:sindri_overview}
\end{figure}

\section{Method}
\label{sec:method}

\method expands a single real-robot demonstration into policy training episodes with varied motions and appearances (\cref{fig:sindri_overview}). The reconstruction agent builds an interactive simulation scene and refines it by comparing the effects of replayed actions with the observed interaction. In this scene, the trajectory generalization agent samples new object layouts, reuses motion segments that remain feasible, and replans those affected by the layout change. For each successful rollout, the multi-view video generator uses simulated depth to synthesize synchronized camera views under different visual conditions. Each visual variant is paired with the actions and robot states from the same rollout.

\subsection{Reconstruction Agent}
\label{sec:asset_reconstruction_v2}

The reconstruction agent builds a high-fidelity simulation scene from a single real-robot demonstration. Given a source episode $\mathcal{E}$ and the robot's URDF $\mathcal{M}$, the agent outputs the complete simulation assets $\{\mathcal{S}^{\star}\}$, reference trajectories with interaction annotations $\{\tau_{\mathrm{ref}}\}$, and camera extrinsics $\Theta_{\mathrm{cam}}$ to support subsequent trajectory generalization:
\begin{equation}
\mathcal{A}_{\mathrm{recon}}\!\left(\mathcal{E};\mathcal{M}\right)
\rightarrow
\left(\{\mathcal{S}^{\star}\},\{\tau_{\mathrm{ref}}\},\Theta_{\mathrm{cam}}\right).
\label{eq:reconstruction_output_v2}
\end{equation}

\noindent\textbf{Calibration and task grounding.}
The source episode $\mathcal{E}=(\ell,\mathcal{V},\mathbf{A})$ comprises the language instruction $\ell$, multi-view videos $\mathcal{V}$, and recorded actions $\mathbf{A}$. Using the proprioceptive robot states at each frame, the agent renders the URDF $\mathcal{M}$ and optimizes the camera extrinsics $\Theta_{\mathrm{cam}}$ by aligning the rendered masks with the observed robot silhouettes. Grounded in $\ell$ and the observed interactions, it then parses task-specific keyframes to identify the manipulated objects, supporting surfaces, and the reference interaction events $\{\tau_{\mathrm{ref}}\}$.

\noindent\textbf{Geometric reconstruction.}
To recover object geometry, the agent samples views from $\mathcal{V}$ that expose complementary object contours. Collectively, these contours constrain the 3D shape across viewpoints, while the known dimensions of the robot provide a metric scale reference. When sufficient parallax is available, the agent triangulates corresponding landmarks across views to estimate 3D positions~\citep{hartley2004multiview}, with the supporting plane further constraining spatial placement. The resulting geometry, articulated structures, materials, and layout are recorded as an executable scene program. Finally, the agent projects source RGB textures onto visible surfaces and renders the scene to verify object boundaries, metric scale, and visual appearance against the original frames.

\noindent\textbf{Physical reconstruction.}
The agent attaches collision geometries aligned with each object's visual coordinate frame and initializes physical parameters. Both are refined by replaying the recorded actions $\mathbf{A}$ in simulation and diagnosing interaction discrepancies against the real demonstration. For instance, upon a failed grasp, the agent compares the commanded gripper aperture against the measured finger motion: a tracking error in free space triggers a diagnostic check of command units, actuator response, and kinematic linkage; conversely, if the fingers reach the target aperture, the agent inspects the clearance between collision meshes and verifies whether the object is successfully lifted. Together with the geometric assets, these calibrated physical models form the simulation assets $\{\mathcal{S}^{\star}\}$.

\noindent\textbf{Review and Revision.}
The agent pinpoints the earliest divergence between the simulated replay and the real demonstration, leveraging the source views and preceding trajectory to localize the erroneous reconstruction components. After each revision, it compares the updated renderings against the source views and re-executes the recorded actions to verify discrepancy resolution. The reconstruction is accepted once the replay faithfully reproduces all task-required events, as summarized in Algorithm~\ref{alg:reconstruction_agent_v2}.

{\centering
\begin{minipage}{0.85\linewidth}
\begin{algorithm}[H]
\caption{Reconstruction agent}
\label{alg:reconstruction_agent_v2}
\renewcommand{\algorithmicrequire}{\textbf{Inputs:}}
\renewcommand{\algorithmicensure}{\textbf{Outputs:}}
\begin{algorithmic}[1]
\Require Episode $\mathcal E=(\text{instruction }\ell,\,\text{video }\mathcal V,\,\text{action }\mathbf A)$, robot URDF $\mathcal M$
\Ensure Simulation assets $\{\mathcal S^\star\}$, references $\{\tau_{\mathrm{ref}}\}$, extrinsics $\Theta_{\mathrm{cam}}$
\State $\Theta_{\mathrm{cam}} \stackrel{\recongptmark}{\gets}
        \Call{Calibrate}{\mathcal E,\mathcal M}$ \Comment{estimate camera extrinsics}
\State $\mathcal T \stackrel{\recongptmark}{\gets} \Call{Understand}{\mathcal E}$ \Comment{identify objects, supports, and task events}
\State $\mathcal F \stackrel{\recongptmark}{\gets}
        \Call{SelectKeyframes}{\mathcal E,\mathcal T}$ \Comment{event keyframes}
\State $P \stackrel{\recongptmark}{\gets}
        \Call{WriteScene}{\mathcal F,\mathcal T,\Theta_{\mathrm{cam}}}$ \Comment{encode geometry, materials, and layout}
\State $S_0 \gets \Call{Execute}{P}$ \Comment{initialize scene}
\State $\mathcal S_{\mathrm{geo}} \gets
        \Call{Register}{S_0,\mathcal F,\Theta_{\mathrm{cam}}}$ \Comment{align 3D scale and object poses}
\State $\mathcal S_{\mathrm{phys}} \gets \Call{AddPhysics}{\mathcal S_{\mathrm{geo}}}$ \Comment{attach collision and physics models}
\State $Q \gets (\Theta_{\mathrm{cam}},\mathcal S_{\mathrm{geo}},\mathcal S_{\mathrm{phys}})$ \Comment{pack current reconstruction state}
\State $\mathcal D \gets \Call{Compare}{Q,\mathcal E}$ \Comment{detect visual discrepancies}
\State $\mathcal H \gets \varnothing$ \Comment{initialize revision history}
\Loop
    \State $a \stackrel{\recongptmark}{\gets} \Call{Review}{\mathcal T,Q,\mathcal D,\mathcal H}$ \Comment{select next refinement decision}
    \If{$a=\textsc{Accept}$}
        \State \Return $\left(\{\mathcal{S}^{\star}\},\{\tau_{\mathrm{ref}}\},\Theta_{\mathrm{cam}}\right)$ \Comment{export finalized assets}
    \ElsIf{$a=\textsc{Stop}$}
        \State \Return \textsc{Unresolved} \Comment{terminate as unresolved}
    \EndIf
    \State $(Q,\mathcal D,\mathcal H) \gets \Call{ProbeAndRevise}{a,Q,\mathcal E,\mathcal D,\mathcal H}$ \Comment{probe and revise assets}
\EndLoop
\end{algorithmic}
\end{algorithm}
\end{minipage}\par
}

\subsection{Trajectory Generalization Agent}
\label{sec:trajectory_generation}

The trajectory generalization agent receives the simulation assets from the reconstruction agent, adapts the reference motions to new sampled object layouts, executes the resulting trajectories in simulation, and retains successful rollouts to form the demonstration dataset $\mathcal{D}_{\mathrm{sim}}$:
\begin{equation}
\mathcal{A}_{\mathrm{gen}}\!\left(
\{\mathcal{S}^{\star}\},\{\tau_{\mathrm{ref}}\},\Theta_{\mathrm{cam}};
N
\right)
\rightarrow
\mathcal{D}_{\mathrm{sim}},
\label{eq:generalization_output_v2}
\end{equation}
where $N$ is the number of candidate layouts, and $\mathcal{D}_{\mathrm{sim}}$ stores the executed trajectory, simulation-rendered RGB-D videos, and task instruction for each retained rollout.

\noindent\textbf{Motion segmentation.}
The reference trajectory is first partitioned into $K$ ordered motion segments $\tau_i$, spanning stages such as grasp acquisition, handover, and release. As in MimicGen~\citep{mandlekar2023mimicgen}, object-relative targets transfer the demonstration to novel layouts. For each segment, boundary poses, gripper states, and object-relative relations are extracted from the reference replay as segment-specific pose targets and interaction annotations $\mathcal{C}_i$. \Cref{fig:supp_motion_segmentation} illustrates these boundaries across three robot platforms.

\noindent\textbf{Layout sampling.}
For objects designated for spatial variation, the agent leverages the initial baseline frame and reconstructed geometry to define a sampling domain $\Omega$ on the support surface. Drawing poses from $\Omega$ and filtering them via static geometric checks yields $N$ candidate layouts $\mathcal{L}_n$, each specifying the initial object poses for subsequent planning.

\noindent\textbf{Trajectory planning.}
For each sampled layout $\mathcal{L}_n$, the agent adapts segment targets to the new object configuration by transforming relative end-effector targets from their reference frames into the world frame. Given the adapted targets and interaction annotations $\mathcal{C}_i^{(n)}$, it reuses segments unaffected by layout changes and invokes cuRobo~\citep{sundaralingam2023curobo} to sequentially plan the remaining segments. For instance, if the block moves while the drawer remains stationary, the agent retains the handle approach and pull provided their paths stay collision-free, replanning only the reach from the post-pull state. Although the in-drawer destination is unchanged, the carry now originates from the new grasp configuration. Finally, adjacent segments are stitched at their shared boundary states.

\noindent\textbf{Simulation Dataset.}
The simulator executes each trajectory continuously across segment boundaries and retains only rollouts that satisfy the task-specific event checks in Appendix C. For each retained rollout $n$, the simulator renders synchronized multi-view RGB-D observations from the recorded scene states $\mathcal{X}^{(n)}$. These observations are paired with the corresponding simulator state $s_{n,t}$ and action $a_{n,t}$ at each time step $t$ to form a sample in the simulated dataset $\mathcal{D}_{\mathrm{sim}}$. Algorithm~\ref{alg:generalization_agent_v2} summarizes the procedure of trajectory generalization.

{\centering
\begin{minipage}{0.85\linewidth}
\begin{algorithm}[H]
\caption{Trajectory generalization agent}
\label{alg:generalization_agent_v2}
\renewcommand{\algorithmicrequire}{\textbf{Inputs:}}
\renewcommand{\algorithmicensure}{\textbf{Outputs:}}
\begin{algorithmic}[1]
\Require Simulation assets $\{\mathcal S^\star\}$, reference trajectories $\{\tau_{\mathrm{ref}}\}$,
         extrinsics $\Theta_{\mathrm{cam}}$, $N$ candidate layouts, instruction $\ell$ 
\Ensure Simulation Dataset $\mathcal D_{\mathrm{sim}}$
\State $\mathcal B \stackrel{\recongptmark}{\gets} \Call{SegmentReferenceMotion}{\{\tau_{\mathrm{ref}}\}}$
       \Comment{\begin{tabular}[t]{@{}l@{}}motion segments $\mathcal B=((\tau_i,\mathcal C_i))_{i=1}^{K}$\end{tabular}}
\State $\Omega \stackrel{\recongptmark}{\gets} \Call{SelectSamplingRegion}{\{\mathcal S^\star\}}$ \Comment{define sampling region $\Omega$}
\State $L \gets \Call{SampleLayouts}{\Omega,\{\mathcal S^\star\},N}$ \Comment{draw N layouts}
\State $\mathcal D_{\mathrm{sim}} \gets \varnothing$ \Comment{initialize dataset}
\ForAll{$\mathcal L_n\in L$}
    \State $x_0^{(n)} \gets (q_{\mathrm{ref}},\mathcal L_n)$ \Comment{set initial robot and object states}
    \For{$i=1,\ldots,K$}
        \State $\mathcal C_i^{(n)} \gets \Call{AdaptTargets}{\mathcal C_i,x_{i-1}^{(n)}}$
               \Comment{target pose transformation $T_{W,\mathrm{ee}}^{(n)}=T_{W,r}^{(n)}T_{r,\mathrm{ee}}$}
        \State $(\hat\tau_i^{(n)},x_i^{(n)}) \gets
               \Call{ReuseOrReplan}{\tau_i,\mathcal C_i^{(n)},x_{i-1}^{(n)}}$
               \Comment{reuse if valid, else cuRobo}
        \State \textbf{skip layout} if $\hat\tau_i^{(n)}=\varnothing$
    \EndFor
    \State $\hat\tau^{(n)} \gets \Call{JoinAtBoundaries}{(\hat\tau_i^{(n)})_{i=1}^{K}}$ \Comment{stitch segments at boundaries}
    \State $\mathcal X^{(n)} \gets \Call{Execute}{\hat\tau^{(n)},\{\mathcal S^\star\},\mathcal L_n}$
           \Comment{robot and object states over time}
    \If{$\Call{VerifyTaskEvents}{\mathcal X^{(n)},\mathcal B}$}
        \State $\mathcal V^{(n)} \gets
               \Call{RenderRGBD}{\{\mathcal S^\star\},\mathcal X^{(n)},\Theta_{\mathrm{cam}}}$ \Comment{render multi-view RGB-D videos}
        \State $\mathcal D_{\mathrm{sim}} \gets \mathcal D_{\mathrm{sim}}\cup
               \{(\hat\tau^{(n)},\mathcal V^{(n)},\ell)\}$
    \EndIf
\EndFor
\State \Return $\mathcal D_{\mathrm{sim}}$
\end{algorithmic}
\end{algorithm}
\end{minipage}\par
}

\subsection{Multi-View Sim2Real Video Generation}
\label{sec:multiview_generation}

For each successful rollout, the video generator uses simulated depth to guide motion and scene structure, with per-view reference images and a shared caption controlling appearance and scene content.
The generated views inherit the rollout's action and state sequence, providing visual variants of the same execution.

\begin{figure}[!htbp]
  \centering
  \includegraphics[width=.95\textwidth]{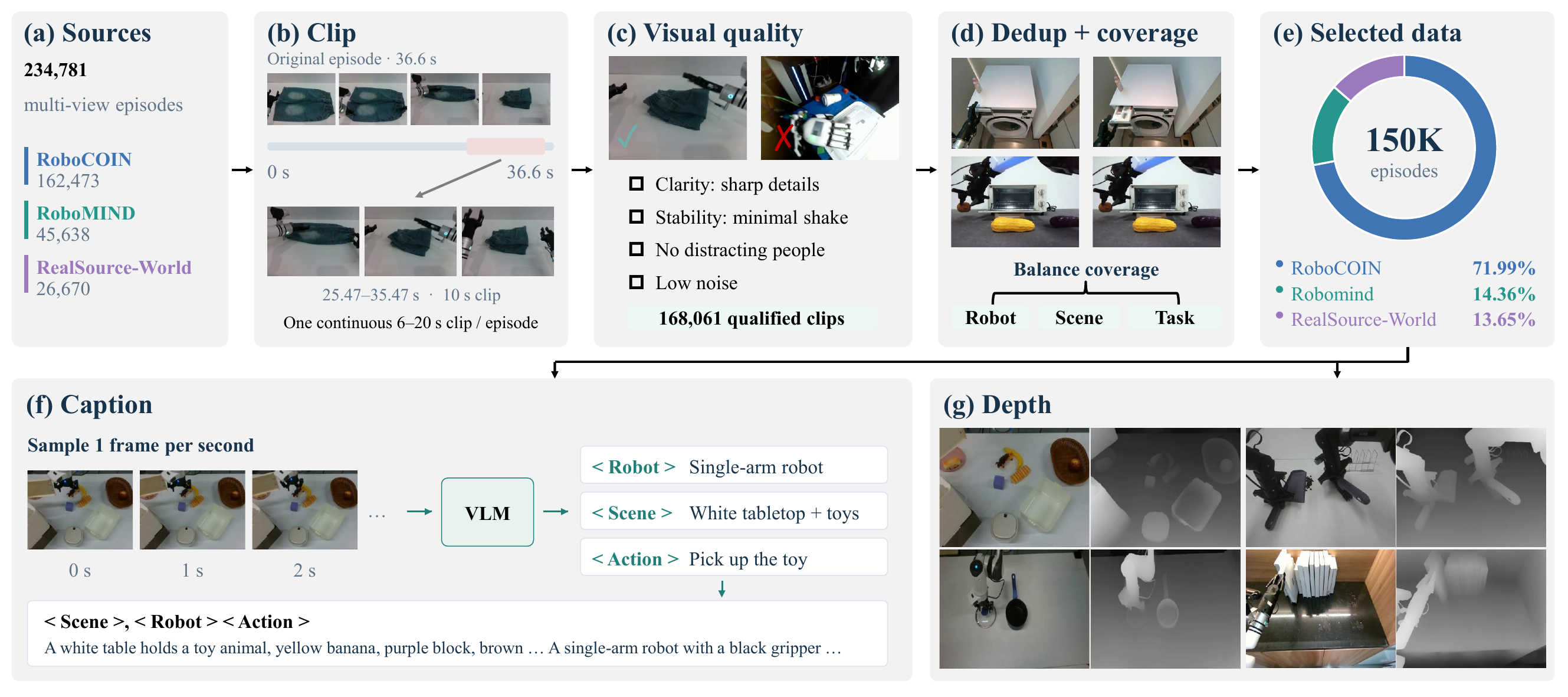}
  \caption{\textbf{Curation of multi-view training clips.} Of 234,781 source episodes, 168,061 pass screening and 155,442 remain after deduplication. Sampling across robot, action, scene, and source yields 150,000 clips, each with a shared caption and per-view depth.}
  \label{fig:supp_curation}
\end{figure}

\noindent\textbf{Training data curation.}
We train the video generator on synchronized real-robot clips. The source inventory comprises 234,781 episodes with at least two views from RoboCOIN~\citep{wu2025robocoin}, RoboMIND~\citep{wu2025robomind}, and RealSource-World~\citep{realman2025realsourceworld}. For each episode we select one continuous 6--20\,s interval, preferably 10\,s, and take the same interval from each camera (\cref{fig:supp_curation}).

Technical checks flag decoding, timing, and image defects. Semantic screening checks that the robot and its manipulation are visible and that the primary view is clear and stable; auxiliary views must decode over the selected interval. Screening retains 168,061 clips.
\FloatBarrier

We merge records by episode identity and exact primary-video hashes, then screen for near duplicates using perceptual hashes at ordered timestamps. From the remaining 155,442 clips, we select 150,000 by cycling through robot, action, scene, and source strata, prioritizing quality within each stratum. RoboCOIN, RoboMIND, and RealSource-World contribute 71.99\%, 14.36\%, and 13.65\%, respectively. 

\par\noindent\begin{minipage}{\linewidth}
  \begin{minipage}[t]{.505\linewidth}
    \vspace{0pt}
    \textbf{Caption and depth controls.}
    Using frames sampled at one frame per second, Qwen3.7-plus~\citep{alibabacloud2026qwen37plus} describes the scene, robot, and action. All cameras and temporal windows share the resulting caption, providing a common semantic description for the full clip. Video Depth Anything Small~\citep{chen2025videodepthanything} estimates relative depth for each camera. We normalize depth over the entire clip, mapping near regions to bright values and far regions to dark values, so the encoding stays consistent across windows. For appearance, we sample one common timestamp and keep its per-camera RGB references fixed throughout the clip, providing a stable appearance anchor for each view.
    
    \smallskip
    \textbf{Shared backbone and controls.}
    All three cameras share the VAE, text encoder, and DiT of Wan2.2-Fun-5B-Control~\citep{alibabapai_wan22fun5bcontrol,peebles2023dit}, while retaining separate video tokens for each view. The DiT receives each camera's encoded depth alongside its noisy video latents. Reference-image tokens form a prefix for within-view self-attention, while text cross-attention supplies the shared caption. At blocks 6, 12, 18, and 24, the three streams exchange features at the same latent time, allowing each view to incorporate complementary information from the other two. 
    % Learned camera-role embeddings distinguish the primary and wrist streams. 
    Within each exchange block, all cameras read the other cameras' features from the same pre-update state, making the exchange independent of stream update order. 
  \end{minipage}\hfill
  \begin{minipage}[t]{.45\linewidth}
    \vspace{0pt}\centering
    \includegraphics[width=\linewidth]{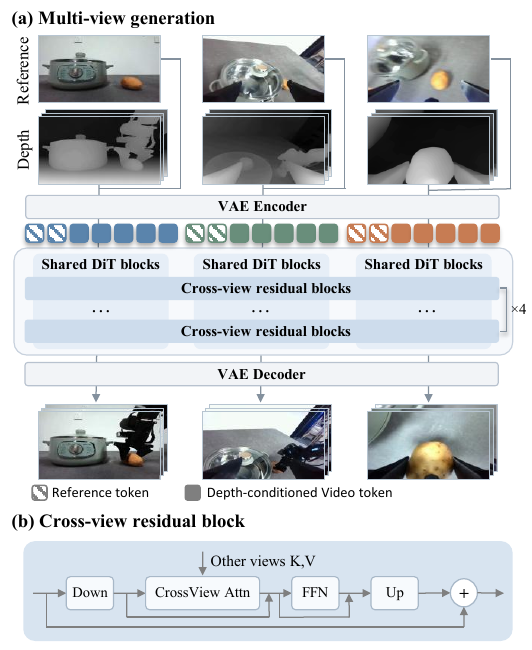}
    \captionsetup{skip=3pt,singlelinecheck=false}
    \captionof{figure}{\textbf{Multi-view video generator.} (a) The camera streams share a DiT with LoRA adapters. (b) Cross-view attention uses queries from one camera and keys and values from the other two.}
    \label{fig:multiview_dit}
  \end{minipage}
\end{minipage}\par

\noindent\textbf{Cross-view residual adaptation.}
Each residual block in \cref{fig:multiview_dit} shares its parameters across views.
Queries come from the target view, with keys and values pooled from the other views at the same latent time.
The bottleneck projects 3,072 channels to 1,024 and uses 16 attention heads. 
At insertion point $\ell$, the video features $H_\ell^{v}$ of view $v$ are updated as
\begin{equation}
 H_\ell^{v,+}=H_\ell^{v}+
 U_\ell\,\mathcal{M}_\ell\!\left(
 D_\ell H_\ell^{v},\{D_\ell H_\ell^{u}\}_{u\ne v}\right),
 \label{eq:cross_view}
\end{equation}
where $D_\ell$ normalizes and projects features into a low-dimensional bottleneck, $\mathcal{M}_\ell$ combines cross-view attention and a feed-forward network with residual connections and diffusion-timestep modulation, and $U_\ell$ restores the original dimension.
Zero-initializing $U_\ell$ makes each block an identity mapping at initialization~\citep{zhang2023controlnet}.

\noindent\textbf{Staged training.}
We train the cross-view blocks for 4K steps, then jointly optimize them with shared rank-32 LoRA adapters~\citep{hu2022lora} in DiT self-attention for another 16K steps, keeping the pretrained backbone frozen.
The first stage learns communication on top of fixed per-view features; the second adapts within-view generation and cross-view communication together.
Both stages use the flow-matching objective~\citep{lipman2023flow}.
Given clean latents $z_0^v$ and independent standard Gaussian noise $\epsilon^v$, all views share a diffusion timestep $t$ and noise level $\sigma=\sigma(t)$ from the base schedule.
For supervised elements, noisy latents and target velocities are
\begin{equation}
 z_\sigma^v=(1-\sigma)z_0^v+\sigma\epsilon^v,
 \qquad u^v=\epsilon^v-z_0^v.
 \label{eq:flow_target}
\end{equation}
For the joint model's conditioned velocity prediction $\hat{u}_\theta^v$, we minimize
\begin{equation}
 \mathcal{L}_{\mathrm{FM}}=\mathbb{E}\!\left[
 \frac{1}{|\mathcal{V}|}\sum_{v\in\mathcal{V}}
 \frac{\|M^v\odot(\hat{u}_\theta^v-u^v)\|_2^2}
 {\|M^v\|_1}
 \right],
 \label{eq:flow_loss}
\end{equation}
where $\mathcal{V}$ indexes available views and $M^v$ selects supervised elements across latent channels, excluding padding and the hard-clamped first latent in history-conditioned samples.
Normalizing by the number of supervised elements gives each available view equal weight in the objective.
The shared LoRA adapters act on the query, key, value, and output projections in all 30 native self-attention blocks. Together with the four cross-view modules, they account for 100.3 million trainable parameters in the second stage.

\noindent\textbf{Long-video generation.}
We train on 121-frame windows at $640\!\times\!352$ and 24 FPS, conditioning approximately 25\% on a 25-frame ground-truth history prefix through the backbone's mask/video pathway.
At inference, each continuation uses the preceding window's last 25 generated frames, hard-clamping the first latent and using the remaining history as soft conditioning.
Reference images remain fixed while depth controls advance along the rollout.
For a partial final window, we repeat the last depth frame and trim the output to the rollout duration.
We discard the overlapping 25 frames and append the remaining 96, preserving action and state alignment.
Appendix~\ref{supp:video_data} lists screening thresholds and deduplication settings, and Appendix~\ref{supp:video_model} details the model architecture and training procedure.

\section{Experiments}
\label{sec:experiments}

Our central question is whether the demonstrations generated by \method can train policies that succeed on real robots. We first examine whether replaying recorded actions in a reconstructed scene reproduces the observed interaction and how often planning succeeds when the object layout changes. We then compare generated camera views with real recordings and ask experts to rank videos generated in new scenes. To connect these evaluations to policy performance, we compare policies trained on real, simulated, or \method data in the default scene. We then test whether varying the appearance of the synthetic training videos improves generalization without additional real demonstrations and examine how the effect varies across tasks and visual conditions.

\subsection{Sim-Ready Asset Reconstruction}
\label{sec:exp_reconstruction}

\begin{figure}[!htbp]
  \centering
  \begin{minipage}[c]{.51\linewidth}
  \centering
  \includegraphics[width=\linewidth]{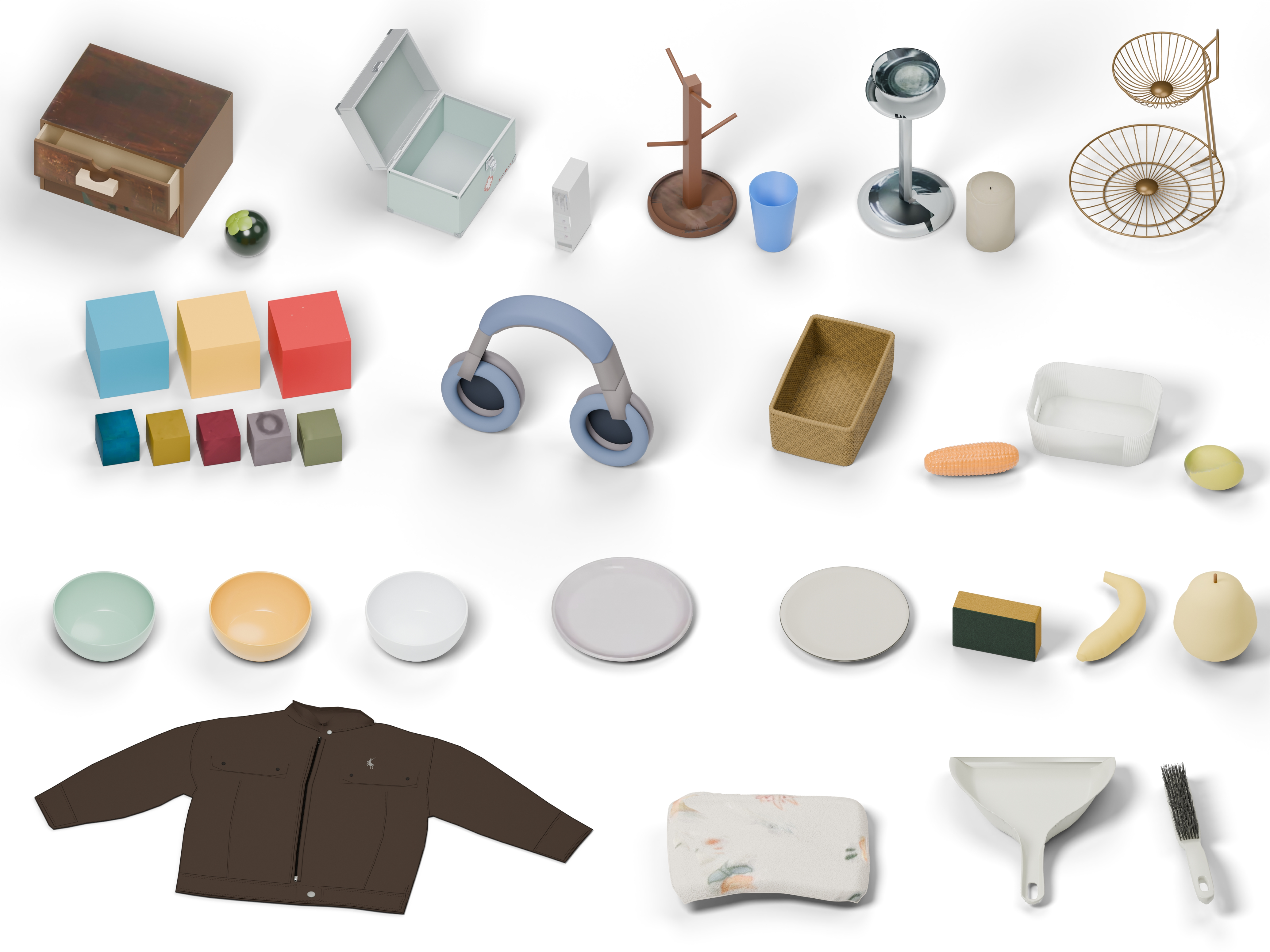}
  \caption{\textbf{Representative reconstructed assets.} Examples cover rigid, articulated, and deformable objects with varied geometry and appearance.}
  \label{fig:reconstruction_cases}
  \end{minipage}\hfill
  \begin{minipage}[c]{.46\linewidth}
  \centering
  \small
  \renewcommand{\arraystretch}{.92}
  \setlength{\tabcolsep}{3pt}
  \captionof{table}{\textbf{Planning success across nine tasks.} Each fraction gives successful plans over sampled layouts. The pooled rate weights each layout equally.}
  \label{tab:supp_planning_results}
  \begin{tabular}{@{}lrr@{}}
    \toprule
    Task & Count & Rate (\%) \\
    \midrule
    \multicolumn{3}{@{}l}{\textit{Astribot}} \\
    Transfer and rehang cup & 80/80 & 100.0 \\
    Place lemon in tray & 80/104 & 76.9 \\
    \multicolumn{3}{@{}l}{\textit{AgileX}} \\
    Place object in drawer & 80/80 & 100.0 \\
    Place candle on stand & 80/80 & 100.0 \\
    Stack three cubes & 80/88 & 90.9 \\
    \multicolumn{3}{@{}l}{\textit{Franka}} \\
    Place banana on plate & 80/91 & 87.9 \\
    \multicolumn{3}{@{}l}{\textit{UR5e}} \\
    Pack medicine and close the lid & 80/85 & 94.1 \\
    \multicolumn{3}{@{}l}{\textit{Tienyi 2.0}} \\
    Place corn in basket & 80/80 & 100.0 \\
    Nest three bowls & 80/80 & 100.0 \\
    \midrule
    Pooled & 720/768 & 93.8 \\
    \bottomrule
  \end{tabular}
  \end{minipage}
\end{figure}

\Cref{fig:reconstruction_cases} illustrates reconstruction across rigid, articulated, and deformable objects, including curved and thin structures, containers, and jointed mechanisms.
Replay exposes errors that are difficult to diagnose from rendered assets alone. In the banana and corn cases, the agent corrected mechanical clearance, checked the gripper command, and realigned collision geometry after a camera update. The gripper change improved the corn interaction, while the banana grasp still failed. In the lemon case, a proposed thickness estimate slightly improved held-out fitting error but worsened alignment in the full head-camera view, so the agent retained the previous geometry. Appendix~\ref{sec:supp_agent_evidence} gives the revision records.

\subsection{Trajectory Generalization}
\label{sec:exp_trajectory_generalization}

We evaluate trajectory generalization on nine tasks across five platforms: Astribot, AgileX, Franka, UR5e, and Tienyi 2.0.
For each task, we sample object layouts and plan until obtaining 80 successful trajectories, then compute planning success rate over all attempts.
Across the nine tasks, \method achieves an overall planning success rate of \textbf{93.8\%} (\cref{tab:supp_planning_results}).
One attempt corresponds to one joint layout, including all relevant objects.The rate measures planning before the execution checks in \cref{sec:trajectory_generation}.
\Cref{fig:supp_planning_layouts} locates these outcomes within each task's sampled region. 

\noindent\begin{minipage}{\linewidth}
  \centering
  \includegraphics[width=1\linewidth]{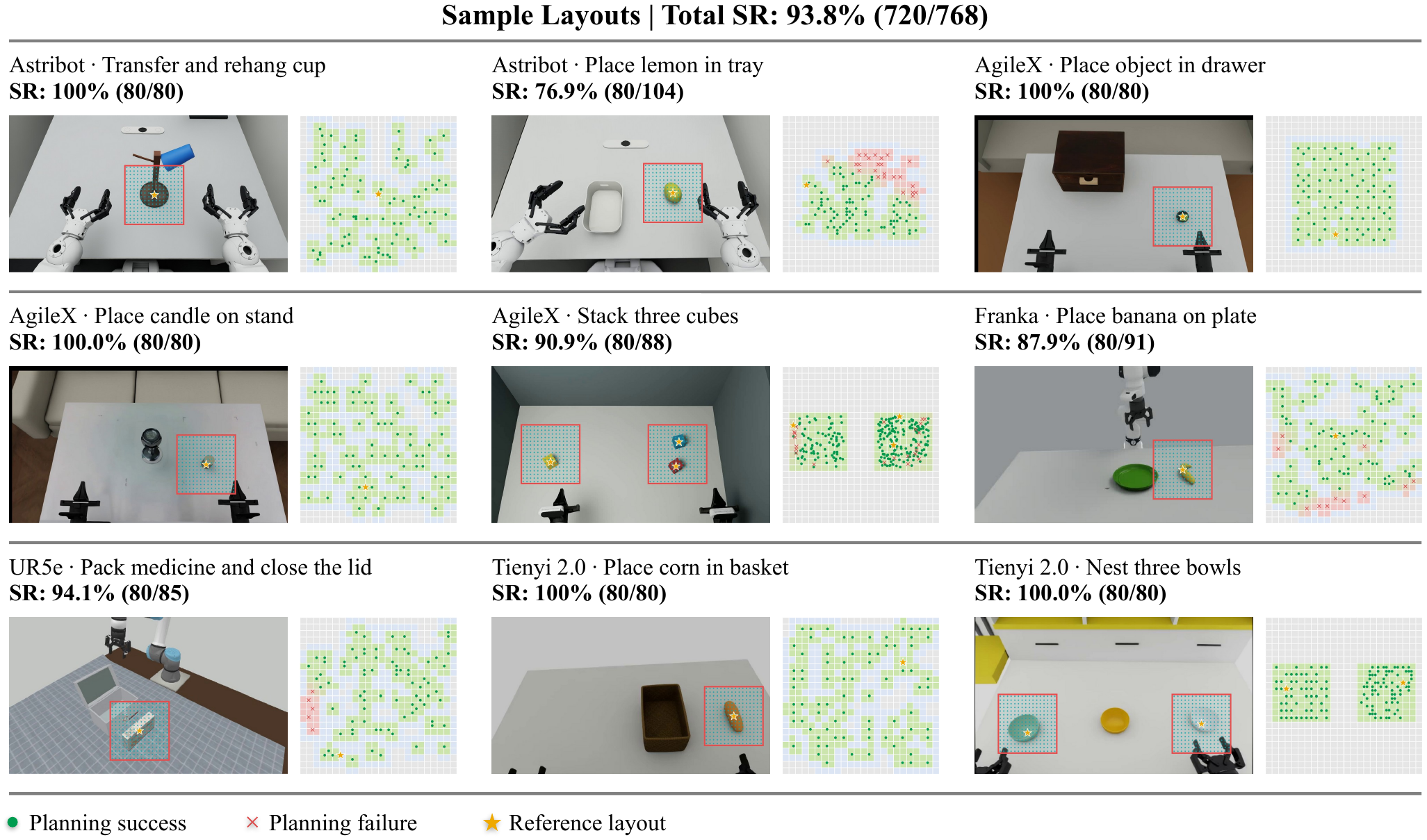}
  \captionof{figure}{\textbf{Spatial distribution of planning outcomes.} For each task, the highlighted regions show where object layouts are sampled. The adjacent plot marks the reference layout (star), successful plans (green dots), and failed plans (red crosses).}
  \label{fig:supp_planning_layouts}
\end{minipage}

\FloatBarrier

\subsection{Multi-View Video Generation}
\label{sec:exp_video_generation}

We compare with RoboRender \citep{huang2026roborender}, Wan-Fun-Control \citep{alibabapai_wan22fun5bcontrol}, and Cosmos-Transfer2.5 \citep{nvidia2025cosmostransfer25}, measuring reconstruction fidelity on real videos and perceptual quality under varied synthetic scene appearances.

\noindent\textbf{Paired real-video evaluation.}
We evaluate 500 held-out episodes with 1,424 matched camera views using FID~\citep{heusel2017fid}, FVD~\citep{unterthiner2018fvd}, PSNR, SSIM~\citep{wang2004ssim}, and LPIPS~\citep{zhang2018lpips}.

\noindent\begin{minipage}{\linewidth}
  \centering
  \includegraphics[width=.96\linewidth]{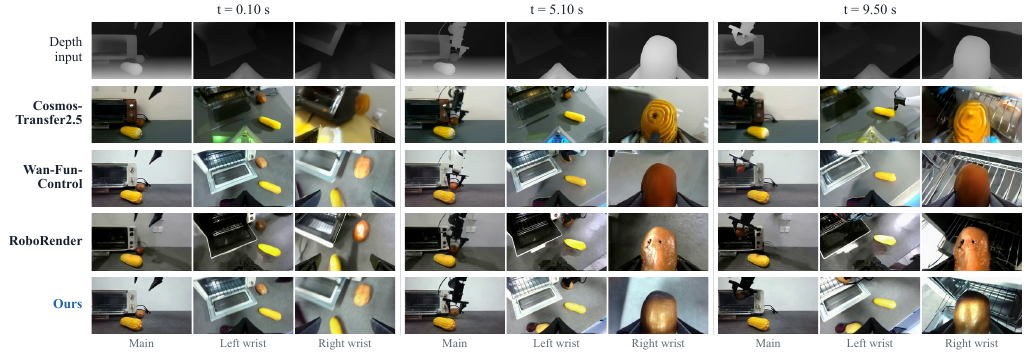}
  \captionof{figure}{\textbf{Qualitative video comparison.} Main and wrist views at three timestamps show scene structure, object appearance, and close-range manipulation generated from depth inputs.}
  \label{fig:video_cases}
\end{minipage}

All 500 episodes contribute a primary view; 462 also contribute two matched wrist views. We align outputs to a common $416\times240$ crop and exclude the initial frame. Image metrics use 32 aligned frames per view; FVD uses a centered 64-frame sequence at 8 FPS. We average PSNR, SSIM, and LPIPS within each view and then each episode. FID and FVD use pooled features.

\begin{table}[!htbp]
    \centering
    \caption{
        \textbf{Video fidelity on held-out real-robot episodes.} Scores compare generated and recorded video across all 1,424 matched views and separately for the 500 primary and 924 wrist views. Bold marks the best score in each block.}
    \label{tab:real_eva_comparison}
    \small
    \setlength{\tabcolsep}{4pt}
    \renewcommand{\arraystretch}{.96}
    \begin{tabular*}{\linewidth}{@{\extracolsep{\fill}}llccccc@{}}
        \toprule
        Subset & Method
        & FID $\downarrow$
        & FVD $\downarrow$
        & PSNR $\uparrow$
        & SSIM $\uparrow$
        & LPIPS $\downarrow$ \\
        \midrule
        All views
        & RoboRender & 23.105 & 272.640 & 14.100 & 0.6068 & 0.4202 \\
        & Cosmos-Transfer2.5 & 65.616 & 906.624 & 10.818 & 0.4091 & 0.5770 \\
        & Wan-Fun-Control & 17.815 & 333.984 & 15.764 & 0.6709 & 0.3136 \\
        & Ours       & \textbf{6.415} & \textbf{65.961}
                     & \textbf{21.340} & \textbf{0.7688}
                     & \textbf{0.1862} \\
        \midrule
        Primary views
        & RoboRender & 35.613 & 382.923 & 14.936 & 0.6158 & 0.3816 \\
        & Cosmos-Transfer2.5 & 66.749 & 934.625 & 12.294 & 0.4589 & 0.5009 \\
        & Wan-Fun-Control & 23.698 & 458.660 & 17.293 & 0.7198 & 0.2144 \\
        & Ours       & \textbf{10.468} & \textbf{104.493}
                     & \textbf{22.510} & \textbf{0.7978}
                     & \textbf{0.1278} \\
        \midrule
        Wrist views
        & RoboRender & 29.246 & 339.961 & 13.349 & 0.5912 & 0.4454 \\
        & Cosmos-Transfer2.5 & 82.730 & 1108.689 & 9.835 & 0.3672 & 0.6274 \\
        & Wan-Fun-Control & 23.473 & 372.287 & 14.463 & 0.6280 & 0.3802 \\
        & Ours       & \textbf{9.025} & \textbf{92.197}
                     & \textbf{20.036} & \textbf{0.7400}
                     & \textbf{0.2284} \\
        \bottomrule
    \end{tabular*}
\end{table}

Conditioning follows each system's generation pipeline. Wan-Fun-Control and \method use real first-frame references, RoboRender uses masked source RGB retaining the robot, and Cosmos uses depth and captions. RoboRender and \method generate views jointly; the other systems generate each view independently.
Across all matched views, our model leads on all five metrics (\cref{tab:real_eva_comparison}), achieving FID 6.415 and FVD 65.961 versus the next-best values of 17.815 and 272.640.
Compared with Wan-Fun-Control, LPIPS falls from 0.2144 to 0.1278 on primary views and from 0.3802 to 0.2284 on wrist views.
% The advantage holds for both primary and wrist views in the supplementary breakdown, covering global scene observations and close-range manipulation views.
The qualitative comparison in \cref{fig:video_cases} shows consistent scene structure and object appearance across views and time.
On all four systems, wrist-view LPIPS remains higher than primary-view LPIPS.

\begin{figure}[t]
  \centering
  \includegraphics[width=.95\linewidth]{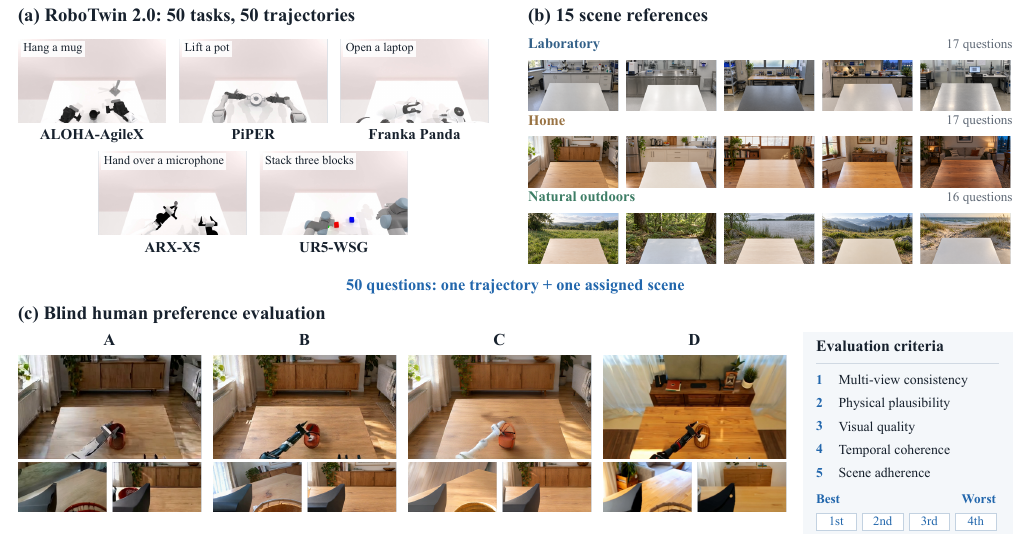}
  \caption{\textbf{Open-scene evaluation of multi-view video generation.} Each of 50 RoboTwin 2.0 trajectories is paired with one of 15 scene references. For each pair, twelve experts independently rank the four anonymized outputs on five criteria.}
  \label{fig:supp_benchmark}
\end{figure}

\clearpage
\noindent\textbf{Open-scene human preference.}
We pair 50 RoboTwin 2.0 task trajectories~\citep{chen2026robotwin2} with scene references spanning laboratory, home, and outdoor environments.
Each trajectory receives one of 15 scene references, yielding 17 laboratory, 17 home, and 16 outdoor task--scene pairs (\cref{fig:supp_benchmark}). Twelve experts rank anonymized, randomly ordered outputs on multi-view consistency, physical plausibility, visual quality, temporal coherence, and task-and-scene adherence.
Each expert ranks all four methods on all 50 pairs, yielding 3,000 rankings across the five criteria. We weight criteria equally and bootstrap evaluators with their complete responses to obtain 95\% intervals for the fixed set of task--scene pairs. The pipelines use different depth encodings: the baselines use fixed camera-specific ranges, with clipping observed in 22 of 100 wrist streams, while \method normalizes inverse depth over each view's sequence.

\begin{table}[!htbp]
  \centering
  \small
  \setlength{\tabcolsep}{3pt}
  \renewcommand{\arraystretch}{1.08}
  \caption{\textbf{Expert rankings on the 50 open-scene task--scene pairs.} Criterion scores are mean ranks from 1 (best) to 4. Overall weights the five criteria equally. Top-1 and Top-2 give the percentage of individual rankings in which a method places first or in the top two.}
  \label{tab:human-preference}
  \begin{tabular*}{\textwidth}{@{\extracolsep{\fill}}l*{8}{c}@{}}
    \toprule
    Method & MVC $\downarrow$ & PP $\downarrow$ & VQ $\downarrow$ & TC $\downarrow$ & SA $\downarrow$ & Overall $\downarrow$ & Top-1 (\%) $\uparrow$ & Top-2 (\%) $\uparrow$ \\
    \midrule
    \textbf{Ours} & \textbf{1.510} & \textbf{1.530} & \textbf{1.438} & \textbf{1.583} & \textbf{1.513} & \textbf{1.515} & \textbf{62.67} & \textbf{88.47} \\
    RoboRender & 2.032 & 2.280 & 2.363 & 2.225 & 2.173 & 2.215 & 23.30 & 64.17 \\
    Wan-Fun-Control & 3.245 & 3.293 & 2.418 & 3.182 & 2.625 & 2.953 & 8.77 & 29.37 \\
    Cosmos-Transfer2.5 & 3.213 & 2.897 & 3.780 & 3.010 & 3.688 & 3.318 & 5.27 & 18.00 \\
    \bottomrule
  \end{tabular*}
  \par\smallskip
  \begin{minipage}{\textwidth}
    \footnotesize
    MVC: multi-view consistency; PP: physical plausibility; VQ: visual quality;
    TC: temporal coherence; SA: task-and-scene adherence.
  \end{minipage}
\end{table}

Our model achieves the best mean rank on each of the five criteria, with an overall mean rank of 1.515 versus 2.215 for RoboRender (\cref{tab:human-preference}). Experts rank our model first in 62.67\% of judgments and among the top two in 88.47\%. Visual quality has the strongest mean rank (1.438), followed by multi-view consistency (1.510); temporal coherence is the weakest of the five for our model (1.583). These judgments complement the paired-video scores by testing whether the same simulated motion remains legible across cameras when the target appearance changes.
Appendix~\ref{supp:open_protocol} details the ranking protocol, and Appendix~\ref{supp:video_cases} shows qualitative comparisons.

\subsection{Real-World Policy Evaluation}
\label{sec:exp_real_world}

We fine-tune $\pi_{0.5}$~\citep{physicalintelligence2025pi05} on four tasks across three platforms: \textit{Lemon Placement} and \textit{Cup Handover} on Astribot, \textit{Drawer Loading} on AgileX, and \textit{Corn Placement} on Tienyi 2.0.
Each policy uses 80 demonstrations per task and the same training configuration: 30K full fine-tuning steps, a batch size of 256, and an action chunk length of 50.
We evaluate 10 real-world rollouts per task, training source, and condition, randomizing object positions within the training-data coverage while keeping task fixtures fixed.
Success requires the full task sequence. For example, the drawer must close after the fruit is placed inside, and the cup handover must end with the receiving arm placing the cup to the left of the stand. Appendix~\ref{supp:real_world_tasks} gives all four completion criteria.

\begin{figure}[t]
  \centering
  \includegraphics[width=1\textwidth]{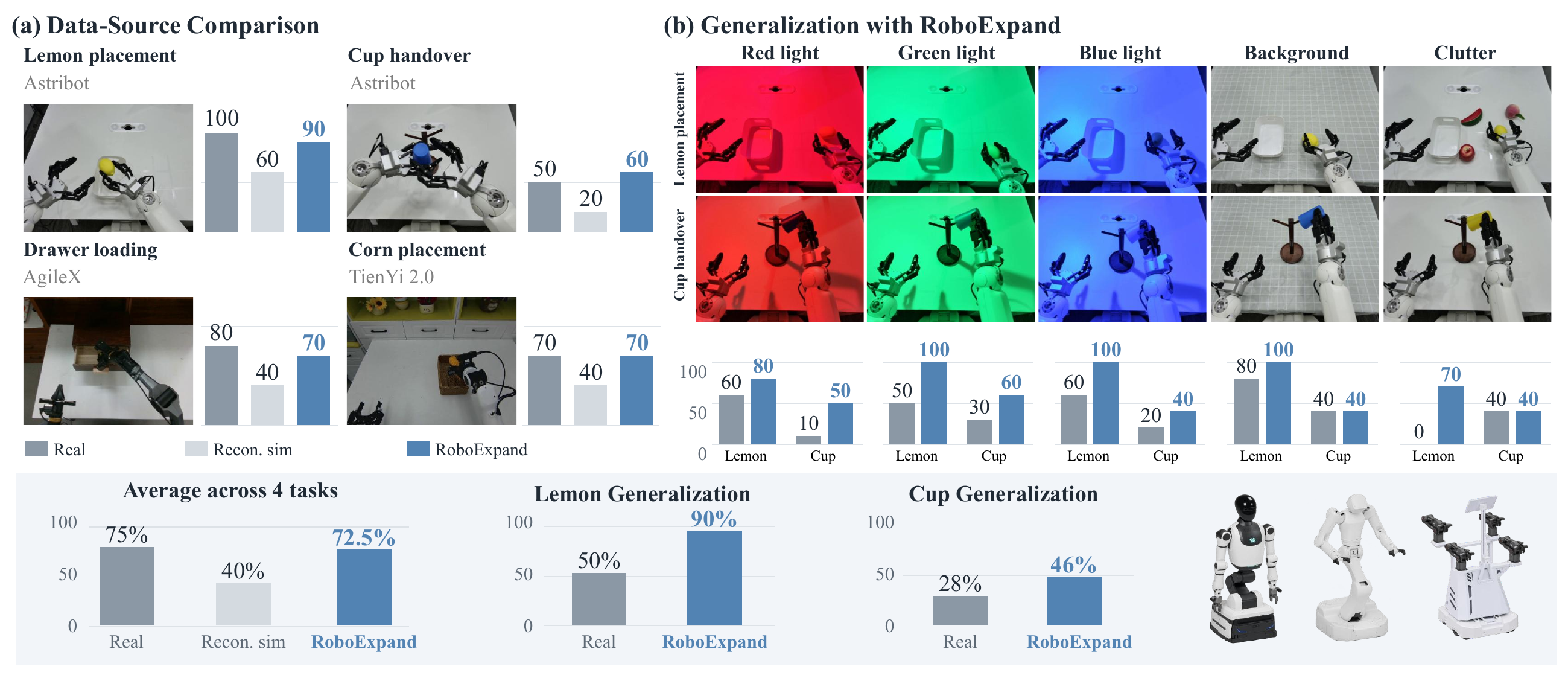}
  \caption{\textbf{Real-world policy performance.} (a) Training-source comparison across four tasks and three platforms. (b) Visual robustness on Astribot: default-scene real data versus synthetic data covering five appearance variations. Each policy uses 80 demonstrations per task and 10 evaluation rollouts per condition.}
  \label{fig:real_world}
\end{figure}

\noindent\textbf{Does \method generate useful data for real-world policy learning?}
We evaluate the complete data engine by training policies on its expanded trajectories and generated observations, and compare with teleoperated real demonstrations and reconstructed simulation data in the default scenes (\cref{fig:real_world}a and \cref{tab:real_world_main}).
\begin{table}[!htbp]
  \centering
  \caption{\textbf{Real-robot task success in the default scene.} For each task, policies train on 80 demonstrations from the indicated source. Each task entry is the percentage of ten physical trials completed successfully.}
  \label{tab:real_world_main}
  \small
  \setlength{\tabcolsep}{3pt}
  \renewcommand{\arraystretch}{1.1}
  \begin{tabularx}{\textwidth}{@{}l*{5}{>{\centering\arraybackslash}X}@{}}
    \toprule
    \multirow{2}{*}{Training data} & \multicolumn{2}{c}{Astribot} & AgileX & Tienyi 2.0 & \multirow{2}{*}{Average} \\
    \cmidrule(lr){2-3}\cmidrule(lr){4-4}\cmidrule(lr){5-5}
    & Lemon placement & Cup handover & Drawer loading & Corn placement & \\
    \midrule
    Real              & \textbf{100} & 50 & \textbf{80} & \textbf{70} & \textbf{75.0} \\
    Reconstructed sim & 60 & 20 & 40 & 40 & 40.0 \\
    RoboExpand        & 90 & \textbf{60} & 70 & \textbf{70} & 72.5 \\
    \bottomrule
  \end{tabularx}
  \par\smallskip
  \begin{minipage}{\textwidth}
    \footnotesize
    Real denotes teleoperated demonstrations. The average weights the four tasks equally.
  \end{minipage}
\end{table}

Across 120 physical trials, policies trained on \method data complete 29/40 trials (72.5\%), compared with 16/40 (40.0\%) for raw simulator renderings and 30/40 (75.0\%) for real demonstrations. The two simulator-derived sources both supply action labels from the reconstructed environment. With 80 demonstrations per task, the policies trained with generated observations complete three or four more trials per task than those trained with raw renderings.
The task-level results span pick-and-place, bimanual coordination, and articulated manipulation: \method matches real-data training on \textit{Corn Placement} and achieves 60\% versus 50\% on \textit{Cup Handover}, while remaining 10 points lower on \textit{Lemon Placement} and \textit{Drawer Loading}.

\noindent\textbf{Can observation expansion support generalization without additional real demonstrations?}
We use \method to extend the two Astribot tasks to varied visual conditions through synthetic observation generation (\cref{fig:real_world}b), following appearance augmentation for robot learning~\citep{tobin2017domainrandomization,yu2025lucidsim}.
Each synthetic training set contains 30 default-scene demonstrations and 10 for each of five conditions: Red light, Green light, Blue light, Background, and Clutter.
Clutter introduces fruit distractors for Lemon Placement and changes cup color for Cup Handover.
We compare this 80-demonstration mixture with 80 default-scene real demonstrations, evaluating each visual condition separately on the physical robot.

Expanded synthetic training achieves 68\% mean success across the ten task--condition pairs, compared with 39\% for default-scene real-data training, a gain of 29 percentage points.
\textit{Lemon Placement} improves from 50\% to 90\%, and \textit{Cup Handover} from 28\% to 46\%.
These averages weight the five tested conditions equally within each task, over 200 physical trials ($2$ tasks $\times$ $2$ sources $\times$ $5$ conditions $\times$ $10$ trials).

\begin{table}[!htbp]
  \centering
  \caption{\textbf{Visual generalization on Astribot.} For each task, policies trained on default-scene real data or a mix of synthetic appearances face the same five test conditions. Each condition reports success over ten physical trials (\%). Average weights the five conditions equally.}
  \label{tab:real_world_generalization}
  \small
  \setlength{\tabcolsep}{3pt}
  \renewcommand{\arraystretch}{1.1}
  \begin{tabularx}{\textwidth}{@{}ll*{6}{>{\centering\arraybackslash}X}@{}}
    \toprule
    \multirow{2}{*}{Task} & \multirow{2}{*}{Training data} & \multicolumn{3}{c}{Light} & \multirow{2}{*}{Background} & \multirow{2}{*}{Clutter} & \multirow{2}{*}{Average} \\
    \cmidrule(lr){3-5}
    & & Red & Green & Blue & & & \\
    \midrule
    Lemon placement & Real & 60 & 50 & 60 & 80 & 0 & 50.0 \\
    & RoboExpand (mix) & \textbf{80} & \textbf{100} & \textbf{100} & \textbf{100} & \textbf{70} & \textbf{90.0} \\
    \midrule
    Cup handover & Real & 10 & 30 & 20 & \textbf{40} & \textbf{40} & 28.0 \\
    & RoboExpand (mix) & \textbf{50} & \textbf{60} & \textbf{40} & \textbf{40} & \textbf{40} & \textbf{46.0} \\
    \bottomrule
  \end{tabularx}
  \par\smallskip
  \begin{minipage}{\textwidth}
    \footnotesize
    Real uses 80 default-scene demonstrations. The synthetic mix uses 30 default-scene demonstrations and 10 per variation. Background changes the tablecloth. Clutter adds fruit distractors for lemon placement and changes the cup color for handover.
  \end{minipage}
\end{table}

\FloatBarrier

\noindent\textbf{How does broader training coverage affect policy robustness?}
\textit{Lemon Placement} improves under all five conditions, reaching 100\% success under green and blue illumination and the changed background.
The largest gain occurs with fruit distractors, from zero to seven successful placements in ten trials.
\textit{Cup Handover} improves under all three lighting conditions by 20--40 percentage points, while background and cup-color performance remain at 40\%.
Across 50 trials per task, the expanded training set yields 20 additional completions for lemon placement and nine for cup handover (\cref{tab:real_world_generalization}). The latter nine all come from the lighting conditions. At the same demonstration budget, the benefit of covering the tested appearances thus depends on both the task and the visual change.

\subsection{Ablation Studies}
\label{sec:exp_ablations}

We compare four video-generator variants trained from the same base checkpoint for 4K updates with matched data and sampling, evaluating all three views of 500 held-out episodes (\cref{tab:multiview_ablation}).
All four variants use the same depth, captions, first-frame references, and per-view sampling noise. Here all three cameras are available for every episode, yielding 1,500 streams and 1,779 synchronized windows per variant. We score full $640\times352$ frames. Final-segment LPIPS covers the newly emitted frames of the last window. Overlap L1 compares the regenerated history with the preceding tail at each continuation, excluding the first, hard-constrained frame.

\begin{table}[!htbp]
    \centering
    \small
    \setlength{\tabcolsep}{4pt}
    \renewcommand{\arraystretch}{1.12}
    \caption{
        \textbf{Effects of multi-view modeling and history conditioning.}
        All variants train for 4K updates and are evaluated for reconstruction fidelity and continuation quality on three views of 500 test episodes.
        History training indicates conditioning on preceding frames during training; all variants use generated history at inference.
        Best results are in \textbf{bold}.
    }
    \label{tab:multiview_ablation}

    \begin{tabularx}{\linewidth}{@{}>{\raggedright\arraybackslash}Xcccccc@{}}
        \toprule
        \multicolumn{2}{c}{Model configuration}
        & \multicolumn{3}{c}{Reconstruction fidelity}
        & \multicolumn{2}{c}{Continuation quality} \\
        \cmidrule(lr){1-2}
        \cmidrule(lr){3-5}
        \cmidrule(lr){6-7}
        Multi-view formulation
        & \shortstack{History\\training}
        & PSNR$\uparrow$
        & SSIM$\uparrow$
        & LPIPS$\downarrow$
        & \shortstack{Final-segment\\LPIPS$\downarrow$}
        & \shortstack{Overlap L1$\downarrow$\\($10^{-2}$)} \\
        \midrule
        Independent views
        & $\checkmark$
        & 17.84
        & 0.7216
        & 0.2846
        & 0.3275
        & 1.664 \\

        Latent concatenation~\citep{huang2026roborender}
        & $\checkmark$
        & 18.29
        & 0.7294
        & 0.2727
        & 0.3181
        & \textbf{1.592} \\

        Cross-view residual modules
        & --
        & 19.09
        & 0.7437
        & 0.2554
        & 0.3057
        & 3.140 \\

        Cross-view residual modules (full)
        & $\checkmark$
        & \textbf{19.33}
        & \textbf{0.7473}
        & \textbf{0.2489}
        & \textbf{0.2934}
        & 1.650 \\
        \bottomrule
    \end{tabularx}
\end{table}

\noindent\textbf{Cross-view residual modules.}
With history training fixed, we compare independent views, spatial latent concatenation adapted from RoboRender~\citep{huang2026roborender}, and residual cross-view modules.
With the same history conditioning, exchanging residual features lowers per-view LPIPS from 0.2846 for independent generation to 0.2489. The improvement persists on newly emitted frames in the last window (0.3275 to 0.2934). This comparison holds training length and conditioning fixed while changing how the cameras share information.
Latent concatenation yields the lowest overlap L1 ($1.592\times10^{-2}$), while its final-segment LPIPS is 0.3181. Agreement on a previously generated prefix and fidelity to real observations on newly generated frames capture different aspects of continuation.

\noindent\textbf{History-conditioned training.}
With the residual architecture fixed, history-conditioned training reduces overlap L1 by 47.5\% and final-segment LPIPS from 0.3057 to 0.2934.
Both variants use generated history at inference; training with history improves overlap agreement and continuation fidelity.
Appendices~\ref{supp:ablation_training} and~\ref{supp:ablation_protocol} detail the ablation training and evaluation settings.

\section{Conclusion}
\label{sec:conclusion}

We presented \method, an agentic Real2Sim2Real data engine that expands a single real-robot demonstration into diverse training experience. 
By coupling action-replay-guided reconstruction and trajectory adaptation with synchronized multi-view video generation, \method expands motion and appearance coverage while retaining aligned action supervision.
Across tasks and robot platforms, synthetic data supports policies that approach real-data training performance, while broader visual coverage improves robustness under appearance variations.
\method supports a practical path to reduce repeated real-robot data collection, reusing a single demonstration to expand training coverage across object layouts and visual conditions.

% We presented \method, an agentic Real2Sim2Real data engine that expands a single real-robot demonstration into diverse trajectories and visual observations for policy learning. 
% The engine first reconstructs an interactive scene, refines it by replaying the recorded actions, and adapts the demonstrated motion to new object layouts. It then uses its learned multi-view transfer model to generate synchronized videos under varied visual conditions, keeping the videos aligned with the trajectory's actions and states.
% Across four tasks on three robots, policies trained on \method-generated data raised average real-world success rate from 40.0\% to 72.5\%, compared with 75.0\% for real-data training at the same budget of 80 demonstrations per task.
% \method substantially lowers two practical barriers to using simulation for real-world policy learning by automating interactive scene reconstruction and narrowing the visual gap through learned video transfer.

% \clearpage

\setlength{\bibsep}{2pt plus .5pt minus .5pt}
\bibliographystyle{plainnat}
\bibliography{custom}

@inproceedings{openx2024,
  author = {{Open X-Embodiment Collaboration}},
  title = {{Open X-Embodiment}: Robotic Learning Datasets and {RT-X} Models},
  booktitle = {IEEE International Conference on Robotics and Automation},
  pages = {6892--6903},
  year = {2024},
  doi = {10.1109/ICRA57147.2024.10611477},
  url = {https://ieeexplore.ieee.org/document/10611477}
}

@inproceedings{khazatsky2024droid,
  author = {Alexander Khazatsky and Karl Pertsch and Suraj Nair and Ashwin Balakrishna and Sudeep Dasari and Siddharth Karamcheti and Soroush Nasiriany and Mohan Kumar Srirama and Lawrence Yunliang Chen and Kirsty Ellis and Peter David Fagan and Joey Hejna and Masha Itkina and Marion Lepert and Yecheng Jason Ma and Patrick Tree Miller and Jimmy Wu and Suneel Belkhale and Shivin Dass and Huy Ha and Arhan Jain and Abraham Lee and Youngwoon Lee and Marius Memmel and Sungjae Park and Ilija Radosavovic and Kaiyuan Wang and Albert Zhan and Kevin Black and Cheng Chi and Kyle Beltran Hatch and Shan Lin and Jingpei Lu and Jean Mercat and Abdul Rehman and Pannag R Sanketi and Archit Sharma and Cody Simpson and Quan Vuong and Homer Rich Walke and Blake Wulfe and Ted Xiao and Jonathan Heewon Yang and Arefeh Yavary and Tony Z. Zhao and Christopher Agia and Rohan Baijal and Mateo Guaman Castro and Daphne Chen and Qiuyu Chen and Trinity Chung and Jaimyn Drake and Ethan Paul Foster and Jensen Gao and David Antonio Herrera and Minho Heo and Kyle Hsu and Jiaheng Hu and Donovon Jackson and Charlotte Le and Yunshuang Li and Roy Lin and Zehan Ma and Abhiram Maddukuri and Suvir Mirchandani and Daniel Morton and Tony Nguyen and Abigail O'Neill and Rosario Scalise and Derick Seale and Victor Son and Stephen Tian and Emi Tran and Andrew E. Wang and Yilin Wu and Annie Xie and Jingyun Yang and Patrick Yin and Yunchu Zhang and Osbert Bastani and Glen Berseth and Jeannette Bohg and Ken Goldberg and Abhinav Gupta and Abhishek Gupta and Dinesh Jayaraman and Joseph J Lim and Jitendra Malik and Roberto Mart{\'i}n-Mart{\'i}n and Subramanian Ramamoorthy and Dorsa Sadigh and Shuran Song and Jiajun Wu and Michael C. Yip and Yuke Zhu and Thomas Kollar and Sergey Levine and Chelsea Finn},
  title = {{DROID}: A Large-Scale In-The-Wild Robot Manipulation Dataset},
  booktitle = {Robotics: Science and Systems},
  year = {2024},
  doi = {10.15607/RSS.2024.XX.120},
  url = {https://www.roboticsproceedings.org/rss20/p120.html}
}

@inproceedings{mandlekar2023mimicgen,
  author = {Ajay Mandlekar and Soroush Nasiriany and Bowen Wen and Iretiayo Akinola and Yashraj Narang and Linxi Fan and Yuke Zhu and Dieter Fox},
  title = {{MimicGen}: A Data Generation System for Scalable Robot Learning using Human Demonstrations},
  booktitle = {Conference on Robot Learning},
  year = {2023},
  volume = {229},
  series = {Proceedings of Machine Learning Research},
  pages = {1820--1864},
  url = {https://proceedings.mlr.press/v229/mandlekar23a.html}
}

@inproceedings{chen2026robotwin2,
  author = {Tianxing Chen and Zanxin Chen and Baijun Chen and Zijian Cai and Yibin Liu and Zixuan Li and Qiwei Liang and Xianliang Lin and Yiheng Ge and Zhenyu Gu and Weiliang Deng and Yubin Guo and Tian Nian and Xuanbing Xie and Qiangyu Chen and Kailun Su and Tianling Xu and Guodong Liu and Mengkang Hu and Huan-Ang Gao and Kaixuan Wang and Zhixuan Liang and Yusen Qin and Xiaokang Yang and Ping Luo and Yao Mu},
  title = {{RoboTwin 2.0}: A Scalable Data Generator and Benchmark with Strong Domain Randomization for Robust Bimanual Robotic Manipulation},
  booktitle = {Proceedings of the 43rd International Conference on Machine Learning},
  series = {Proceedings of Machine Learning Research},
  volume = {306},
  pages = {13673--13699},
  year = {2026},
  url = {https://proceedings.mlr.press/v306/chen26j.html}
}

@inproceedings{yu2025real2render2real,
  author = {Justin Yu and Letian Fu and Huang Huang and Karim El-Refai and Rares Andrei Ambrus and Richard Cheng and Muhammad Zubair Irshad and Ken Goldberg},
  title = {{Real2Render2Real}: Scaling Robot Data Without Dynamics Simulation or Robot Hardware},
  booktitle = {Conference on Robot Learning},
  year = {2025},
  volume = {305},
  series = {Proceedings of Machine Learning Research},
  pages = {547--577},
  url = {https://proceedings.mlr.press/v305/yu25a.html}
}

@article{zhang2026autorsr,
  author = {Jiawei Zhang and Chang Liu and Litao Deng and Tao Zhang and Qin Jin},
  title = {{AutoRSR}: Automated Real-to-Sim-to-Real Data Generation for Scalable Robot Learning},
  journal = {arXiv preprint arXiv:2610.20336},
  year = {2026},
  url = {https://arxiv.org/abs/2610.20336}
}

@inproceedings{dai2025acdc,
  author = {Dai, Tianyuan and Wong, Josiah and Jiang, Yunfan and Wang, Chen and Gokmen, Cem and Zhang, Ruohan and Wu, Jiajun and Fei-Fei, Li},
  title = {Automated Creation of Digital Cousins for Robust Policy Learning},
  booktitle = {Proceedings of The 8th Conference on Robot Learning},
  series = {Proceedings of Machine Learning Research},
  volume = {270},
  pages = {4912--4943},
  year = {2025},
  url = {https://proceedings.mlr.press/v270/dai25a.html}
}

@inproceedings{le2025articulateanything,
  title = {{Articulate-Anything}: Automatic Modeling of Articulated Objects via a Vision-Language Foundation Model},
  author = {Le, Long and Xie, Jason and Liang, William and Wang, Hung-Ju and Yang, Yue and Ma, Yecheng Jason and Vedder, Kyle and Krishna, Arjun and Jayaraman, Dinesh and Eaton, Eric},
  booktitle = {International Conference on Learning Representations},
  year = {2025},
  url = {https://arxiv.org/abs/2410.13882}
}

@inproceedings{cao2026physxanything,
  title = {{PhysX-Anything}: Simulation-Ready Physical {3D} Assets from Single Image},
  author = {Cao, Ziang and Hong, Fangzhou and Chen, Zhaoxi and Pan, Liang and Liu, Ziwei},
  booktitle = {Proceedings of the IEEE/CVF Conference on Computer Vision and Pattern Recognition},
  year = {2026},
  url = {https://arxiv.org/abs/2511.13648}
}

@inproceedings{xia2026sage,
  title = {{SAGE}: Scalable Agentic {3D} Scene Generation for Embodied {AI}},
  author = {Xia, Hongchi and Li, Xuan and Li, Zhaoshuo and Ma, Qianli and Xu, Jiashu and Liu, Ming-Yu and Cui, Yin and Lin, Tsung-Yi and Ma, Wei-Chiu and Wang, Shenlong and Song, Shuran and Wei, Fangyin},
  booktitle = {Proceedings of the IEEE/CVF Conference on Computer Vision and Pattern Recognition},
  pages = {22358--22368},
  year = {2026},
  url = {https://openaccess.thecvf.com/content/CVPR2026/html/Xia_SAGE_Scalable_Agentic_3D_Scene_Generation_for_Embodied_AI_CVPR_2026_paper.html}
}

@article{xie2026codimrecon,
  title = {{CoDimRecon}: Agentic Reconstruction of Sim-Ready {3D} Scenes with Deformable Curves, Surfaces, and Volumes},
  author = {Xie, Shuzhao and Wang, Lelin and Lin, Guying and Wang, Zhi and Li, Minchen},
  journal = {arXiv preprint arXiv:2609.36024},
  year = {2026},
  url = {https://arxiv.org/abs/2609.36024}
}

@article{jang2025dreamgen,
  title = {{DreamGen}: Unlocking Generalization in Robot Learning through Video World Models},
  author = {Joel Jang and Seonghyeon Ye and Zongyu Lin and Jiannan Xiang and Johan Bjorck and Yu Fang and Fengyuan Hu and Spencer Huang and Kaushil Kundalia and Yen-Chen Lin and Loic Magne and Ajay Mandlekar and Avnish Narayan and You Liang Tan and Guanzhi Wang and Jing Wang and Qi Wang and Yinzhen Xu and Xiaohui Zeng and Kaiyuan Zheng and Ruijie Zheng and Ming-Yu Liu and Luke Zettlemoyer and Dieter Fox and Jan Kautz and Scott Reed and Yuke Zhu and Linxi Fan},
  journal = {arXiv preprint arXiv:2505.12705},
  year = {2025},
  url = {https://arxiv.org/abs/2505.12705}
}

@inproceedings{fang2025rebot,
  author = {Fang, Yu and Yang, Yue and Zhu, Xinghao and Zheng, Kaiyuan and Bertasius, Gedas and Szafir, Daniel and Ding, Mingyu},
  title = {{ReBot}: Scaling Robot Learning with Real-to-Sim-to-Real Robotic Video Synthesis},
  booktitle = {IEEE/RSJ International Conference on Intelligent Robots and Systems},
  year = {2025},
  doi = {10.1109/IROS60139.2025.11246305},
  url = {https://ieeexplore.ieee.org/document/11246305}
}

@article{nvidia2025cosmostransfer,
  title = {{Cosmos-Transfer1}: Conditional World Generation with Adaptive Multimodal Control},
  author = {{NVIDIA} and Hassan Abu Alhaija and Jose Alvarez and Maciej Bala and Tiffany Cai and Tianshi Cao and Liz Cha and Joshua Chen and Mike Chen and Francesco Ferroni and Sanja Fidler and Dieter Fox and Yunhao Ge and Jinwei Gu and Ali Hassani and Michael Isaev and Pooya Jannaty and Shiyi Lan and Tobias Lasser and Huan Ling and Ming-Yu Liu and Xian Liu and Yifan Lu and Alice Luo and Qianli Ma and Hanzi Mao and Fabio Ramos and Xuanchi Ren and Tianchang Shen and Xinglong Sun and Shitao Tang and Ting-Chun Wang and Jay Wu and Jiashu Xu and Stella Xu and Kevin Xie and Yuchong Ye and Xiaodong Yang and Xiaohui Zeng and Yu Zeng},
  journal = {arXiv preprint arXiv:2503.14492},
  year = {2025},
  url = {https://arxiv.org/abs/2503.14492}
}

@article{liu2025robotransfer,
  title = {{RoboTransfer}: Controllable Geometry-Consistent Video Diffusion for Manipulation Policy Transfer},
  author = {Liu Liu and Xiaofeng Wang and Guosheng Zhao and Keyu Li and Wenkang Qin and Jiagang Zhu and Jiaxiong Qiu and Zheng Zhu and Guan Huang and Zhizhong Su},
  journal = {arXiv preprint arXiv:2505.23171},
  year = {2025},
  url = {https://arxiv.org/abs/2505.23171}
}

@article{yang2025orv,
  title = {{ORV}: {4D} Occupancy-centric Robot Video Generation},
  author = {Xiuyu Yang and Bohan Li and Shaocong Xu and Nan Wang and Chongjie Ye and Zhaoxi Chen and Minghan Qin and Yikang Ding and Zheng Zhu and Xin Jin and Hang Zhao and Hao Zhao},
  journal = {arXiv preprint arXiv:2506.03079},
  year = {2025},
  url = {https://arxiv.org/abs/2506.03079}
}

@article{wang2026robovip,
  title = {{RoboVIP}: Multi-View Video Generation with Visual Identity Prompting Augments Robot Manipulation},
  author = {Boyang Wang and Haoran Zhang and Shujie Zhang and Jinkun Hao and Mingda Jia and Qi Lv and Yucheng Mao and Zhaoyang Lyu and Jia Zeng and Xudong Xu and Jiangmiao Pang},
  journal = {arXiv preprint arXiv:2601.05241},
  year = {2026},
  url = {https://arxiv.org/abs/2601.05241}
}

@article{huang2026roborender,
  title = {{RoboRender}: Robot-Oriented Video Generation for Visual Sim-to-Real Transfer},
  author = {Huang Huang and Wensi Ai and Ziyu Chen and Youhui Wang and Zijian Du and Yang Liu and Jiaolong Yang and Li Fei-Fei and Jiajun Wu},
  journal = {arXiv preprint arXiv:2610.09254},
  year = {2026},
  url = {https://arxiv.org/abs/2610.09254}
}

@inproceedings{torne2024rialto,
  author = {Marcel Torne Villasevil and Anthony Simeonov and Zechu Li and April Chan and Tao Chen and Abhishek Gupta and Pulkit Agrawal},
  title = {Reconciling Reality through Simulation: A Real-To-Sim-to-Real Approach for Robust Manipulation},
  booktitle = {Proceedings of Robotics: Science and Systems},
  year = {2024},
  address = {Delft, Netherlands},
  month = jul,
  doi = {10.15607/RSS.2024.XX.015},
  url = {https://www.roboticsproceedings.org/rss20/p015.html}
}

@inproceedings{wu2025robomind,
  author = {Kun Wu and Chengkai Hou and Jiaming Liu and Zhengping Che and Xiaozhu Ju and Zhuqin Yang and Meng Li and Yinuo Zhao and Zhiyuan Xu and Guang Yang and Shichao Fan and Xinhua Wang and Fei Liao and Zhen Zhao and Guangyu Li and Zhao Jin and Lecheng Wang and Jilei Mao and Ning Liu and Pei Ren and Qiang Zhang and Yaoxu Lyu and Mengzhen Liu and Jingyang He and Yulin Luo and Zeyu Gao and Chenxuan Li and Chenyang Gu and Yankai Fu and Di Wu and Xingyu Wang and Sixiang Chen and Zhenyu Wang and Pengju An and Siyuan Qian and Shanghang Zhang and Jian Tang},
  title = {{RoboMIND}: Benchmark on Multi-embodiment Intelligence Normative Data for Robot Manipulation},
  booktitle = {Proceedings of Robotics: Science and Systems},
  year = {2025},
  doi = {10.15607/RSS.2025.XXI.152},
  url = {https://www.roboticsproceedings.org/rss21/p152.html}
}

@article{wu2025robocoin,
  author = {Shihan Wu and Xuecheng Liu and Shaoxuan Xie and Pengwei Wang and Xinghang Li and Bowen Yang and Zhe Li and Kai Zhu and Hongyu Wu and Yiheng Liu and Zhaoye Long and Runtian Xu and Yue Wang and Chong Liu and Dihan Wang and Ziqiang Ni and Xiang Yang and You Liu and Ruoxuan Feng and Lei Zhang and Denghang Huang and Chenghao Jin and Anlan Yin and Xinlong Wang and Zhenguo Sun and Junkai Zhao and Mengfei Du and Mingyu Cao and Xiansheng Chen and Hongyang Cheng and Xiaojie Zhang and Yankai Fu and Ning Chen and Cheng Chi and Sixiang Chen and Huaihai Lyu and Xiaoshuai Hao and Yequan Wang and Bo Lei and Dong Liu and Xi Yang and Yance Jiao and Tengfei Pan and Yunyan Zhang and Songjing Wang and Ziqian Zhang and Xu Liu and Ji Zhang and Caowei Meng and Zhizheng Zhang and Jiyang Gao and Song Wang and Xiaokun Leng and Zhiqiang Xie and Zhenzhen Zhou and Peng Huang and Wu Yang and Yandong Guo and Yichao Zhu and Suibing Zheng and Hao Cheng and Xinmin Ding and Yang Yue and Huanqian Wang and Chi Chen and Jingrui Pang and YuXi Qian and Haoran Geng and Lianli Gao and Haiyuan Li and Bin Fang and Gao Huang and Yaodong Yang and Hao Dong and He Wang and Hang Zhao and Yadong Mu and Di Hu and Hao Zhao and Tiejun Huang and Shanghang Zhang and Yonghua Lin and Zhongyuan Wang and Guocai Yao},
  title = {{RoboCOIN}: An Open-Sourced Bimanual Robotic Data Collection for Integrated Manipulation},
  journal = {arXiv preprint arXiv:2511.17441},
  year = {2025},
  url = {https://arxiv.org/abs/2511.17441}
}

@misc{realman2025realsourceworld,
  author = {{RealMan Robotics}},
  title = {{RealSource World}},
  howpublished = {Hugging Face dataset card},
  year = {2025},
  note = {Accessed October 25, 2026},
  url = {https://huggingface.co/datasets/RealmanRobot/RealSource-World}
}

@misc{alibabacloud2026qwen37plus,
  author = {{Alibaba Cloud}},
  title = {{Qwen3.7-Plus}},
  howpublished = {Alibaba Cloud Model Studio documentation},
  year = {2026},
  note = {Accessed November 2, 2026},
  url = {https://www.alibabacloud.com/help/en/model-studio/qwen3-7-plus}
}

@misc{alibabapai_wan22fun5bcontrol,
  author = {{Alibaba-PAI}},
  title = {{Wan2.2-Fun-5B-Control}},
  howpublished = {Hugging Face model card},
  year = {n.d.},
  note = {Accessed October 25, 2026},
  url = {https://huggingface.co/alibaba-pai/Wan2.2-Fun-5B-Control}
}

@misc{nvidia2025cosmostransfer25,
  author = {{NVIDIA}},
  title = {{Cosmos-Transfer2.5-2B}},
  howpublished = {Hugging Face model card},
  year = {2025},
  note = {Released October 6, 2025; accessed October 25, 2026},
  url = {https://huggingface.co/nvidia/Cosmos-Transfer2.5-2B}
}

@inproceedings{chen2025videodepthanything,
  author = {Sili Chen and Hengkai Guo and Shengnan Zhu and Feihu Zhang and Zilong Huang and Jiashi Feng and Bingyi Kang},
  title = {{Video Depth Anything}: Consistent Depth Estimation for Super-Long Videos},
  booktitle = {Proceedings of the IEEE/CVF Conference on Computer Vision and Pattern Recognition},
  year = {2025},
  pages = {22831--22840},
  url = {https://openaccess.thecvf.com/content/CVPR2025/html/Chen_Video_Depth_Anything_Consistent_Depth_Estimation_for_Super-Long_Videos_CVPR_2025_paper.html}
}

@inproceedings{hu2022lora,
  title = {{LoRA}: Low-Rank Adaptation of Large Language Models},
  author = {Edward J. Hu and Yelong Shen and Phillip Wallis and Zeyuan Allen-Zhu and Yuanzhi Li and Shean Wang and Lu Wang and Weizhu Chen},
  booktitle = {International Conference on Learning Representations},
  year = {2022},
  url = {https://openreview.net/forum?id=nZeVKeeFYf9}
}

@inproceedings{heusel2017fid,
  author = {Martin Heusel and Hubert Ramsauer and Thomas Unterthiner and Bernhard Nessler and Sepp Hochreiter},
  title = {{GANs} Trained by a Two Time-Scale Update Rule Converge to a Local Nash Equilibrium},
  booktitle = {Advances in Neural Information Processing Systems},
  volume = {30},
  year = {2017},
  url = {https://proceedings.neurips.cc/paper/2017/hash/8a1d694707eb0fefe65871369074926d-Abstract.html}
}

@article{unterthiner2018fvd,
  author = {Thomas Unterthiner and Sjoerd van Steenkiste and Karol Kurach and Raphael Marinier and Marcin Michalski and Sylvain Gelly},
  title = {Towards Accurate Generative Models of Video: A New Metric \& Challenges},
  journal = {arXiv preprint arXiv:1812.01717},
  year = {2018},
  url = {https://arxiv.org/abs/1812.01717}
}

@article{wang2004ssim,
  author = {Zhou Wang and Alan C. Bovik and Hamid R. Sheikh and Eero P. Simoncelli},
  title = {Image Quality Assessment: From Error Visibility to Structural Similarity},
  journal = {IEEE Transactions on Image Processing},
  volume = {13},
  number = {4},
  year = {2004},
  url = {https://ece.uwaterloo.ca/~z70wang/publications/ssim.html}
}

@inproceedings{zhang2018lpips,
  author = {Richard Zhang and Phillip Isola and Alexei A. Efros and Eli Shechtman and Oliver Wang},
  title = {The Unreasonable Effectiveness of Deep Features as a Perceptual Metric},
  booktitle = {Proceedings of the IEEE Conference on Computer Vision and Pattern Recognition},
  year = {2018},
  pages = {586--595},
  url = {https://openaccess.thecvf.com/content_cvpr_2018/html/Zhang_The_Unreasonable_Effectiveness_CVPR_2018_paper.html}
}

@inproceedings{lipman2023flow,
  author = {Yaron Lipman and Ricky T. Q. Chen and Heli Ben-Hamu and Maximilian Nickel and Matthew Le},
  title = {Flow Matching for Generative Modeling},
  booktitle = {The Eleventh International Conference on Learning Representations},
  year = {2023},
  url = {https://openreview.net/forum?id=PqvMRDCJT9t}
}

@article{carion2025sam3,
  author = {Nicolas Carion and Laura Gustafson and Yuan-Ting Hu and Shoubhik Debnath and Ronghang Hu and Didac Suris and Chaitanya Ryali and Kalyan Vasudev Alwala and Haitham Khedr and Andrew Huang and Jie Lei and Tengyu Ma and Baishan Guo and Arpit Kalla and Markus Marks and Joseph Greer and Meng Wang and Peize Sun and Roman R{\"a}dle and Triantafyllos Afouras and Effrosyni Mavroudi and Katherine Xu and Tsung-Han Wu and Yu Zhou and Liliane Momeni and Rishi Hazra and Shuangrui Ding and Sagar Vaze and Francois Porcher and Feng Li and Siyuan Li and Aishwarya Kamath and Ho Kei Cheng and Piotr Doll{\'a}r and Nikhila Ravi and Kate Saenko and Pengchuan Zhang and Christoph Feichtenhofer},
  title = {{SAM 3}: Segment Anything with Concepts},
  journal = {arXiv preprint arXiv:2511.16719},
  year = {2025},
  url = {https://arxiv.org/abs/2511.16719}
}

@inproceedings{skorokhodov2022styleganv,
  author = {Ivan Skorokhodov and Sergey Tulyakov and Mohamed Elhoseiny},
  title = {{StyleGAN-V}: A Continuous Video Generator with the Price, Image Quality and Perks of {StyleGAN2}},
  booktitle = {Proceedings of the IEEE/CVF Conference on Computer Vision and Pattern Recognition},
  pages = {3626--3636},
  year = {2022},
  url = {https://skor.sh/stylegan-v/}
}

@inproceedings{sundaralingam2023curobo,
  author = {Balakumar Sundaralingam and Siva Kumar Sastry Hari and Adam Fishman and Caelan Garrett and Karl Van Wyk and Valts Blukis and Alexander Millane and Helen Oleynikova and Ankur Handa and Fabio Ramos and Nathan Ratliff and Dieter Fox},
  title = {{CuRobo}: Parallelized Collision-Free Robot Motion Generation},
  booktitle = {IEEE International Conference on Robotics and Automation (ICRA)},
  pages = {8112--8119},
  year = {2023},
  doi = {10.1109/ICRA48891.2023.10160765}
}

@inproceedings{xue2025demogen,
  author = {Zhengrong Xue and Shuying Deng and Zhenyang Chen and Yixuan Wang and Zhecheng Yuan and Huazhe Xu},
  title = {{DemoGen}: Synthetic Demonstration Generation for Data-Efficient Visuomotor Policy Learning},
  booktitle = {Proceedings of Robotics: Science and Systems},
  year = {2025},
  doi = {10.15607/RSS.2025.XXI.157},
  url = {https://www.roboticsproceedings.org/rss21/p157.html}
}

@inproceedings{yu2025lucidsim,
  author = {Alan Yu and Ge Yang and Ran Choi and Yajvan Ravan and John Leonard and Phillip Isola},
  title = {Learning Visual Parkour from Generated Images},
  booktitle = {Proceedings of The 8th Conference on Robot Learning},
  series = {Proceedings of Machine Learning Research},
  volume = {270},
  pages = {2500--2516},
  year = {2025},
  url = {https://proceedings.mlr.press/v270/yu25b.html}
}

@inproceedings{xiang2025trellis,
  author = {Jianfeng Xiang and Zelong Lv and Sicheng Xu and Yu Deng and Ruicheng Wang and Bowen Zhang and Dong Chen and Xin Tong and Jiaolong Yang},
  title = {Structured {3D} Latents for Scalable and Versatile {3D} Generation},
  booktitle = {Proceedings of the IEEE/CVF Conference on Computer Vision and Pattern Recognition},
  year = {2025},
  pages = {21469--21480},
  url = {https://openaccess.thecvf.com/content/CVPR2025/html/Xiang_Structured_3D_Latents_for_Scalable_and_Versatile_3D_Generation_CVPR_2025_paper.html}
}

@inproceedings{chen2024urdformer,
  author = {Qiuyu Chen and Aaron Walsman and Marius Memmel and Kaichun Mo and Alex Fang and Dieter Fox and Abhishek Gupta},
  title = {{URDFormer}: A Pipeline for Constructing Articulated Simulation Environments from Real-World Images},
  booktitle = {Proceedings of Robotics: Science and Systems},
  year = {2024},
  doi = {10.15607/RSS.2024.XX.124},
  url = {https://www.roboticsproceedings.org/rss20/p124.html}
}

@article{physicalintelligence2025pi05,
  author = {{Physical Intelligence} and Kevin Black and Noah Brown and James Darpinian and Karan Dhabalia and Danny Driess and Adnan Esmail and Michael Equi and Chelsea Finn and Niccolo Fusai and Manuel Y. Galliker and Dibya Ghosh and Lachy Groom and Karol Hausman and Brian Ichter and Szymon Jakubczak and Tim Jones and Liyiming Ke and Devin LeBlanc and Sergey Levine and Adrian Li-Bell and Mohith Mothukuri and Suraj Nair and Karl Pertsch and Allen Z. Ren and Lucy Xiaoyang Shi and Laura Smith and Jost Tobias Springenberg and Kyle Stachowicz and James Tanner and Quan Vuong and Homer Walke and Anna Walling and Haohuan Wang and Lili Yu and Ury Zhilinsky},
  title = {{$\pi_{0.5}$}: a Vision-Language-Action Model with Open-World Generalization},
  journal = {arXiv preprint arXiv:2504.16054},
  year = {2025},
  url = {https://arxiv.org/abs/2504.16054}
}

@inproceedings{mandlekar2022robomimic,
  title = {What Matters in Learning from Offline Human Demonstrations for Robot Manipulation},
  author = {Mandlekar, Ajay and Xu, Danfei and Wong, Josiah and Nasiriany, Soroush and Wang, Chen and Kulkarni, Rohun and Fei-Fei, Li and Savarese, Silvio and Zhu, Yuke and Mart{\'i}n-Mart{\'i}n, Roberto},
  booktitle = {Proceedings of the 5th Conference on Robot Learning},
  series = {Proceedings of Machine Learning Research},
  volume = {164}, pages = {1678--1690}, year = {2022},
  url = {https://proceedings.mlr.press/v164/mandlekar22a.html}
}

@inproceedings{tobin2017domainrandomization,
  title = {Domain Randomization for Transferring Deep Neural Networks from Simulation to the Real World},
  author = {Josh Tobin and Rachel Fong and Alex Ray and Jonas Schneider and Wojciech Zaremba and Pieter Abbeel},
  booktitle = {IEEE/RSJ International Conference on Intelligent Robots and Systems},
  year = {2017}, doi = {10.1109/IROS.2017.8202133},
  url = {https://arxiv.org/abs/1703.06907}
}

@inproceedings{xiang2020sapien,
  title = {{SAPIEN}: A {SimulAted Part-Based Interactive ENvironment}},
  author = {Fanbo Xiang and Yuzhe Qin and Kaichun Mo and Yikuan Xia and Hao Zhu and Fangchen Liu and Minghua Liu and Hanxiao Jiang and Yifu Yuan and He Wang and Li Yi and Angel X. Chang and Leonidas J. Guibas and Hao Su},
  booktitle = {Proceedings of the IEEE/CVF Conference on Computer Vision and Pattern Recognition},
  year = {2020},
  url = {https://openaccess.thecvf.com/content_CVPR_2020/html/Xiang_SAPIEN_A_SimulAted_Part-Based_Interactive_ENvironment_CVPR_2020_paper.html}
}

@book{hartley2004multiview,
  title = {Multiple View Geometry in Computer Vision},
  author = {Richard Hartley and Andrew Zisserman},
  publisher = {Cambridge University Press}, edition = {2nd}, year = {2004},
  url = {https://www.robots.ox.ac.uk/~vgg/hzbook/}
}

@inproceedings{peebles2023dit,
  title = {Scalable Diffusion Models with Transformers},
  author = {William Peebles and Saining Xie},
  booktitle = {Proceedings of the IEEE/CVF International Conference on Computer Vision},
  year = {2023}, pages = {4195--4205},
  url = {https://openaccess.thecvf.com/content/ICCV2023/html/Peebles_Scalable_Diffusion_Models_with_Transformers_ICCV_2023_paper.html}
}

@article{chung2023unimax,
  title = {{UniMax}: Fairer and more Effective Language Sampling for Large-Scale Multilingual Pretraining},
  author = {Hyung Won Chung and Noah Constant and Xavier Garcia and Adam Roberts and Yi Tay and Sharan Narang and Orhan Firat},
  journal = {arXiv preprint arXiv:2304.09151}, year = {2023},
  url = {https://arxiv.org/abs/2304.09151}
}

@inproceedings{zhang2023controlnet,
  title = {Adding Conditional Control to Text-to-Image Diffusion Models},
  author = {Lvmin Zhang and Anyi Rao and Maneesh Agrawala},
  booktitle = {Proceedings of the IEEE/CVF International Conference on Computer Vision},
  year = {2023}, pages = {3836--3847},
  url = {https://openaccess.thecvf.com/content/ICCV2023/html/Zhang_Adding_Conditional_Control_to_Text-to-Image_Diffusion_Models_ICCV_2023_paper.html}
}

@inproceedings{shi2024mvdream,
  title = {{MVDream}: Multi-view Diffusion for {3D} Generation},
  author = {Yichun Shi and Peng Wang and Jianglong Ye and Long Mai and Kejie Li and Xiao Yang},
  booktitle = {International Conference on Learning Representations}, year = {2024},
  url = {https://proceedings.iclr.cc/paper_files/paper/2024/hash/adbe936993aa7cf41e45054d8b72f183-Abstract-Conference.html}
}

@inproceedings{loshchilov2019adamw,
  title = {Decoupled Weight Decay Regularization},
  author = {Ilya Loshchilov and Frank Hutter},
  booktitle = {International Conference on Learning Representations}, year = {2019},
  url = {https://arxiv.org/abs/1711.05101}
}

@article{ho2022cfg,
  title = {Classifier-Free Diffusion Guidance},
  author = {Jonathan Ho and Tim Salimans},
  journal = {arXiv preprint arXiv:2207.12598}, year = {2022},
  url = {https://arxiv.org/abs/2207.12598}
}

@inproceedings{carreira2017i3d,
  title = {Quo Vadis, Action Recognition? A New Model and the Kinetics Dataset},
  author = {Jo{\~a}o Carreira and Andrew Zisserman},
  booktitle = {Proceedings of the IEEE Conference on Computer Vision and Pattern Recognition},
  year = {2017}, doi = {10.1109/CVPR.2017.502},
  url = {https://openaccess.thecvf.com/content_cvpr_2017/html/Carreira_Quo_Vadis_Action_CVPR_2017_paper.html}
}

@article{efron1979bootstrap,
  title = {Bootstrap Methods: Another Look at the Jackknife},
  author = {Bradley Efron}, journal = {The Annals of Statistics},
  volume = {7}, number = {1}, pages = {1--26}, year = {1979},
  doi = {10.1214/aos/1176344552}
}

@inproceedings{vla-trace,
    title = "{VLA}-Trace: Diagnosing Vision-Language-Action Models through Representation and Behavior Tracing",
    author = "Shi, Haoyuan  and
      Ren, Xiancong  and
      Zhang, Yingji  and
      Zhang, Qinfan  and
      Hu, Jiayu  and
      Shan, Haozhe  and
      Dong, Han  and
      Lu, Jinpeng  and
      Chen, Yinda  and
      Zhang, Yi  and
      Dai, Yong  and
      Ju, Xiaozhu",
    editor = "Lee, Hung-Yi  and
      Dev, Sunipa  and
      Li, Junyi Jessy  and
      Elliott, Desmond",
    booktitle = "Proceedings of the 2026 Conference on Empirical Methods in Natural Language Processing",
    month = oct,
    year = "2026",
    address = "Budapest, Hungary",
    publisher = "Association for Computational Linguistics",
    url = "https://aclanthology.org/2026.emnlp-main.2076/",
    doi = "10.18653/2026.emnlp-main.2076",
    pages = "47280--47315",
    ISBN = "979-8-89176-459-0"
}

@misc{zhang2026geomemvlainternalizing3dscene,
      title={GeoMem-VLA: Internalizing 3D Scene Understanding into Temporal Memory for Vision-Language-Action Models}, 
      author={Qinfan Zhang and Yifan Luo and Haoyuan Shi and Guanlin Liu and Han Dong and Haozhe Shan and Yong Dai and Xiancong Ren and Xiaozhu Ju},
      year={2026},
      eprint={2610.29866},
      archivePrefix={arXiv},
      primaryClass={cs.CV},
      url={https://arxiv.org/abs/2610.29866}, 
}

@misc{zhu2026dream4actsharedvisualaction,
      title={Dream4ACT: A Shared Visual Action Interface for Multi-Embodiment Video-Action Modeling}, 
      author={Xiangyu Zhu and Jin Xu and Yue Guo and Xin Wu and Yifan Sun and Xiancong Ren and Jianxin Sun and Yong Dai and Xiaozhu Ju},
      year={2026},
      eprint={2609.40153},
      archivePrefix={arXiv},
      primaryClass={cs.RO},
      url={https://arxiv.org/abs/2609.40153}, 
}

\clearpage
\appendix
\setcounter{figure}{0}
\setcounter{table}{0}
\setcounter{equation}{0}
\renewcommand{\thefigure}{S\arabic{figure}}
\renewcommand{\thetable}{S\arabic{table}}
\renewcommand{\theequation}{S\arabic{equation}}
\renewcommand{\theHfigure}{supp.\arabic{figure}}
\renewcommand{\theHtable}{supp.\arabic{table}}
\renewcommand{\theHequation}{supp.\arabic{equation}}

\section*{Appendix}

Appendices~\ref{sec:supp_agent_reconstruction}--\ref{sec:supp_agent_evidence} provide additional asset comparisons, motion segmentation examples, rollout acceptance criteria, and revision records. Appendices~\ref{supp:video_data}--\ref{supp:video_cases} document video training and evaluation. Appendix~\ref{supp:real_world_tasks} gives the real-robot evaluation protocol, and Appendix~\ref{supp:dataset_rollout_cases} illustrates generated training episodes and physical policy rollouts.

\section{Additional Reconstructed Assets}
\label{sec:supp_agent_reconstruction}

\begin{figure}[!htbp]
  \centering
  \includegraphics[width=\textwidth]{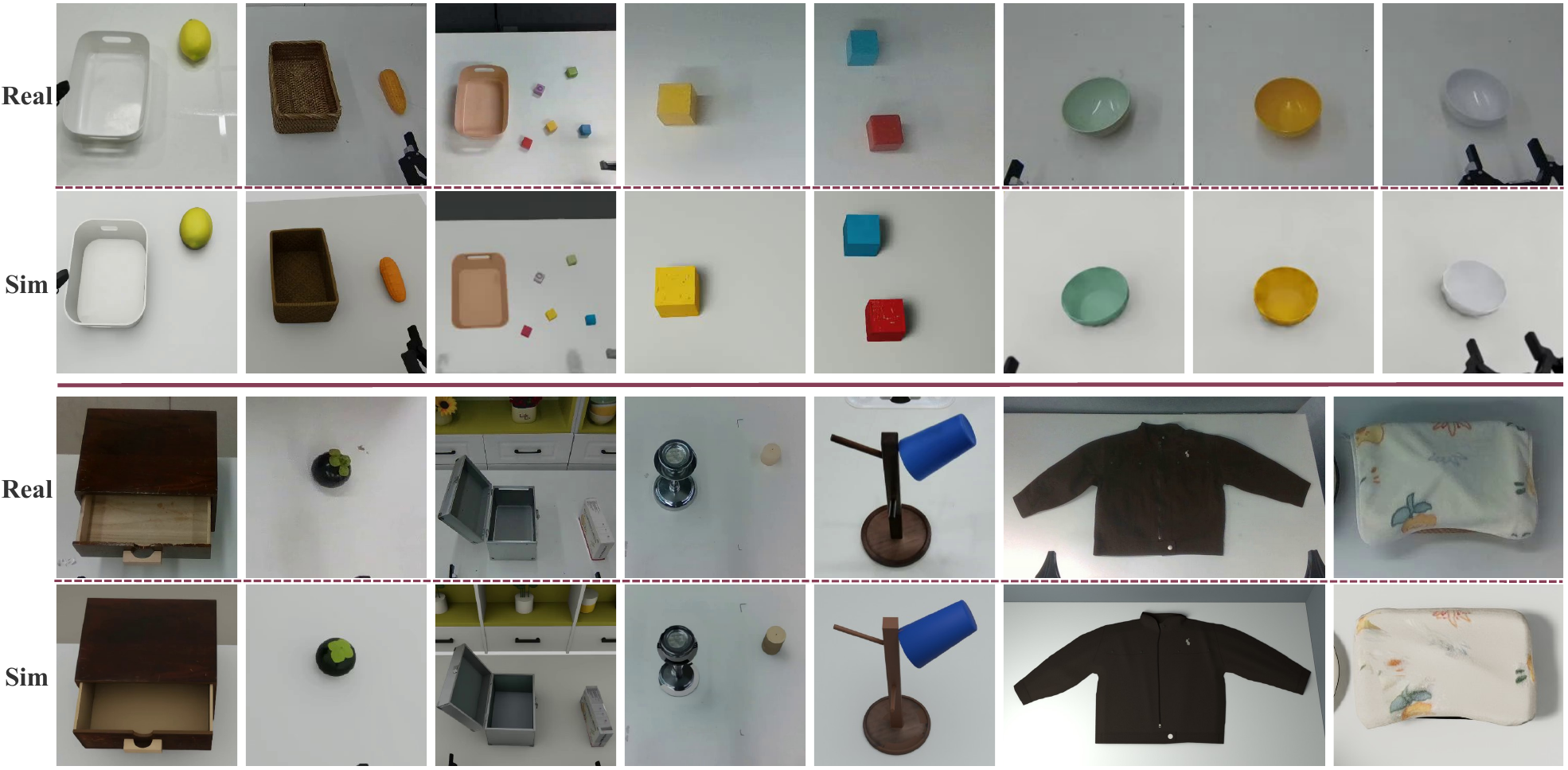}
  \caption{\textbf{Real observations and reconstructed assets.} Each pair of rows compares real observations (Real) with simulation renderings (Sim). Examples include fruits, containers, cubes, bowls, a drawer, a hinged box, a candle and stand, a cup and stand, and deformable objects. These comparisons illustrate reconstructed geometry and appearance; physical behavior is checked separately through action replay.}
  \label{fig:supp_reconstructed_assets}
\end{figure}

Given real-robot demonstrations, our reconstruction agent synthesizes simulation scenes that faithfully capture the geometry, metric scale, spatial layout, and appearance of target objects. The resulting assets span rigid, articulated, and deformable categories across diverse manipulation tasks.
\Cref{fig:supp_reconstructed_assets} presents qualitative comparisons between the source observations and reconstructed assets across scenes.

\section{Motion Segmentation}
\label{sec:supp_motion_segmentation}

The motion segmentation module partitions the reference trajectory $\tau_{\mathrm{ref}}$ into $K$ modular segments $\{\tau_i\}_{i=1}^{K}$ that can be adapted to novel object layouts. Starting from high-level interaction events---such as grasp acquisition, handover, and release---the agent inspects gripper states and object kinematics before and after each event, consulting multi-view observations whenever state transitions are ambiguous. For each confirmed boundary, it logs the event type, reference timestamp, end-effector pose, and gripper state.

Each segment $\tau_i$ corresponds to an atomic manipulation primitive bounded by well-defined start and terminal poses. Boundaries are placed wherever the task transitions to a distinct action phase or a contact-state change necessitates a new motion target. Specifically, the agent expresses each segment's target pose in its governing object's local frame ($T_{r,\mathrm{ee}}$) alongside interaction annotations $\mathcal{C}_i$, enabling layout-invariant segments to be directly reused while affected segments are sequentially replanned via cuRobo without intermediate waypoints.

\begin{figure}[H]
  \centering
  \includegraphics[width=\textwidth]{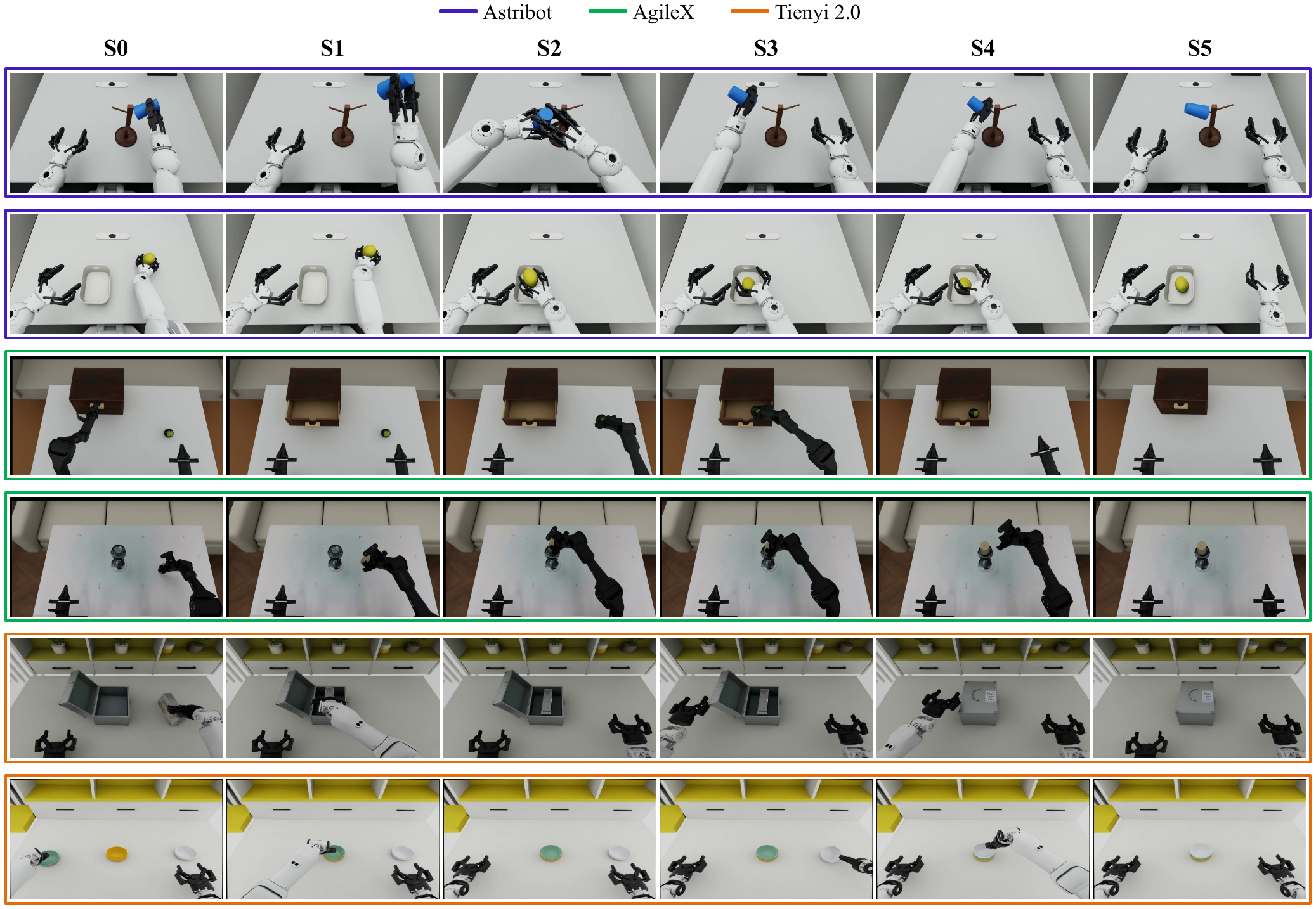}
  \caption{\textbf{Segmenting demonstrations for layout changes.} Read each row from left to right (S0--S5). Changes in the target object or contact phase define the motion segments used for adaptation. Border colors distinguish Astribot, AgileX, and Tienyi 2.0.}
  \label{fig:supp_motion_segmentation}
\end{figure}

\Cref{fig:supp_motion_segmentation} illustrates the resulting segmentation boundaries (S0--S5) across six manipulation tasks on three robot platforms: Astribot, AgileX, and Tienyi 2.0. 

\section{Rollout Acceptance Criteria}
\label{sec:supp_agent_trajectory}

\subsection{Continuous Execution and Event Checks}
\label{supp:rollout_checks}
The simulator executes the assembled trajectory continuously so that contacts and object motion carry from one segment into the next. The task definition specifies the target region, event order, and any timing or stability tolerances. For pick-and-place, acceptance requires all five events in \cref{tab:supp_pick_place_events} in sequence. In the drawer task, the drawer must also be open for placement, and the released block must remain inside through closure.

\begin{table}[!htbp]
  \centering
  \small
  \caption{\textbf{Acceptance criteria for simulated pick-and-place rollouts.} A rollout is accepted only when all five events occur in order within a continuous simulation trace.}
  \label{tab:supp_pick_place_events}
  \begin{tabular}{@{}l@{\hspace{0.8em}}p{0.73\linewidth}@{}}
    \toprule
    Event & Evidence from execution \\
    \midrule
    Grip & Gripper contacts and target motion show that the fingers hold the object. \\
    Lift & The held object leaves its original support. \\
    Carry & The object travels with the gripper to the placement region. \\
    Release & Gripper--object contact ends within the task's release window. \\
    Rest & The object settles in the specified target region. \\
    \bottomrule
  \end{tabular}
\end{table}

\noindent\textbf{Measuring release.}
In the lemon-and-basket case, an early drop exposed a timing error in the release check. Searching for the first contact-free sample \emph{within} the release window can assign an in-window release time to a lemon dropped earlier. The revised check locates the end of the grasp contact in the full trace and compares that time with the prescribed window. An early drop therefore fails the release condition even if the gripper opens on schedule.

\section{Evaluation and Revision Records}
\label{sec:supp_agent_evidence}

\subsection{Revision Protocol}
We define task events and success criteria before evaluating a replay. The agent checks rendered appearance, camera calibration, robot tracking, contacts, and task completion separately. Starting from the first discrepancy, it selects a diagnostic test, revises the relevant asset or control setting, and repeats the affected checks under the same conditions. The record retains both improvements and regressions. Image comparisons assess registration and appearance, while continuous action replay checks contact and object motion.

\subsection{Case Studies}
\noindent\textbf{Banana and corn.}
The agent first corrected mechanical clearance, then checked how the recorded gripper signal related to the required motion. After updating camera calibration, it rechecked object registration and aligned the collision geometry with the registered object. The same gripper change improved the corn task, while the banana grasp still failed.

\noindent\textbf{Cup and stand.}
Early cup motion prompted checks of the cup and stand geometry and of support before grasp. To investigate the subsequent grasp, the agent compared commanded and measured motion, read back drive units, and examined arm tracking separately from finger contact. Paired checks before and after revisions identified improvements under the recorded conditions.

\noindent\textbf{Lemon and basket.}
The agent examined object size and support, gripper linkage and force, and a transport failure following geometry refinement. A candidate thickness from wrist views slightly reduced the held-out fitting error but worsened agreement in the full head-camera view, so the agent retained the previous scene. A separate correction to the release check located the end of grasp contact in the full trace before comparing it with the prescribed release window. Rescoring the same trace then identified an early drop (\cref{supp:rollout_checks}).

\subsection{Rechecking Adopted Revisions}
The agent records adopted asset versions and repeats downstream checks affected by each revision. Camera updates require renewed registration and texture checks; changes to collision geometry also require contact replay. If an evaluation rule changes, both successful and failed traces are rescored from the same recorded states and contacts, with the evaluator version tracked separately.

\section{Video Training Data}
\label{supp:video_data}

\subsection{Quality Screening}
The primary view undergoes full decoding and technical and semantic checks. Auxiliary views must decode throughout the same interval and correspond to distinct physical video files. A clip enters selection only when both technical and semantic screening pass.

\noindent\textbf{Quality checks.}
Pixel-quality metrics use grayscale frames resized to at most 320 pixels wide with aspect ratio preserved; frame identity uses native-resolution RGB. \Cref{tab:supp_quality_thresholds} lists the thresholds. A review flag holds a clip out of the eligible pool until resolved. Low motion alone is recorded as a diagnostic.

\begin{table}[!htbp]
\centering
\small
\setlength{\tabcolsep}{4pt}
\caption{\textbf{Technical screening thresholds.} Intensities are in $[0,255]$; temporal summaries are computed over the selected primary-view interval. $\Delta$ denotes the median frame period.}
\label{tab:supp_quality_thresholds}
\begin{tabularx}{\textwidth}{@{}l X l@{}}
\toprule
Check & Trigger & Outcome \\
\midrule
Blank frames & Either black ($\leq4$) or white ($\geq251$) pixels occupy $\geq99.5\%$ of a frame; such frames occupy $\geq5\%$ of the clip or persist for $\geq0.5$\,s. & Reject \\
Blur & Median Laplacian variance $<6$, or $\geq35\%$ of frames have variance $<4$. & Review \\
Low contrast & Median per-frame grayscale standard deviation $<6$. & Review \\
Impulse noise & For each frame, count pixels whose difference from a $3\times3$ median-filtered image exceeds 45 and whose filtered intensity lies in $(20,235)$; the 90th percentile of this fraction exceeds $5\%$. & Review \\
Frozen video & Consecutive identical native RGB frames persist for $\geq3$\,s and repeated frames comprise $\geq95\%$ of the clip. & Review \\
Timestamp gaps & Maximum frame gap $>\max(0.5\,\mathrm{s},3\Delta)$; smaller gaps exceeding $3\Delta$ are held for review. & Fail / review \\
Low motion & Mean inter-frame grayscale absolute difference $<0.15$. & Flag only \\
\bottomrule
\end{tabularx}
\end{table}

Camera shake is measured every 0.2\,s using Lucas--Kanade corner tracking and a RANSAC partial-affine fit. With at least 12 valid estimates, a clip is flagged for review when all three conditions hold: median normalized translation acceleration $>1.5$, direction-reversal fraction $>0.55$, and median normalized speed $>0.05$. Translation is normalized by image width and height.

\noindent\textbf{Semantic scores.}
Each interval is represented by 20 frames and scored from 0 (unusable) to 4 (excellent). The minimum scores are 3 for clarity, 2 for lighting, 3 for background cleanliness, 3 for action visibility, and 3 for motion stability. Every dimension must pass, with model-reported confidence $\geq0.75$ and an unambiguous visible robot and action. Evidence must cover at least three distinct frames, including clear manipulation at two distinct timestamps.
Confident, frame-supported findings of severe blur, noise, corruption, distracting people, occlusion, or excessive jitter cause rejection; insufficient scores, evidence, or confidence cause review.

The main screening pass uses Qwen3.5-plus, followed by Qwen3.7-plus for all review cases and accepted clips with clarity $\leq3$, stability $\leq3$, or confidence $<0.9$. A deterministic 5\% sample of the remaining accepted clips is also checked by the second model. Conflicting judgments about a visible defect remain under review. The supplementary screening pass uses Qwen3.7-plus alone with the same score floors. Caption generation follows selection (\cref{sec:multiview_generation}).

\subsection{Deduplication and Coverage Selection}
\noindent\textbf{Exact duplicates.}
We merge records with the same original-episode identity, provenance, or source-record identity. We also compare SHA-256 hashes of the full selected primary-view RGB sequence, including every native-resolution frame, frame dimensions, microsecond-relative timestamps, interval duration, and frame count. Among duplicates, we retain the clip with the highest minimum semantic score, then the highest total score, with an ID hash breaking ties deterministically.

\noindent\textbf{Near-duplicate candidates.}
We compute a 64-bit perceptual hash (pHash) from a $32\times32$ grayscale image: apply the DCT, retain the upper-left $8\times8$ coefficients, threshold at the median excluding DC, and set the DC bit to zero. Each clip stores up to 13 uniformly spaced hashes; comparison uses four ordered anchors near the start, one-third, two-thirds, and end. Each hash is divided into four 16-bit bands. A pair is retrieved when the same band matches at all four corresponding anchors.
For anchor-wise Hamming distances $d_1,\ldots,d_4$, a pair is flagged when
\begin{equation}
\max_j d_j\leq3 \quad\text{and}\quad \sum_{j=1}^{4}d_j\leq4.
\end{equation}
Clips whose maximum adjacent-anchor distance is $\leq2$ are omitted from this comparison because their temporal signature is nearly static. Both members of a flagged pair are held out pending identity verification. Search is bounded to 128 records per bucket, 64 candidates per record, two million comparisons, and 100,000 flagged pairs.

\noindent\textbf{Coverage selection.}
Eligible clips are selected by round-robin strata over robot family, action, scene, and source, ranked within strata by minimum semantic score and then total score. The soft source targets are 70\% RoboCOIN, 20\% RoboMIND, and 10\% RealSource-World; unused source allocations are redistributed among eligible clips. The realized source shares appear in \cref{sec:multiview_generation}.

\noindent\textbf{Camera slots.}
The selected cameras occupy primary, left, and right slots. The primary camera may be head-mounted, chest-mounted, or external; auxiliary slots use the designated cameras for each embodiment. For example, 18,711 Franka episodes map their camera-left and camera-right streams to the auxiliary slots. One training clip lacks the right stream; its missing tokens and loss terms are masked, and the loss averages over available views.

\begin{figure}[H]
  \centering
  \includegraphics[width=\textwidth]{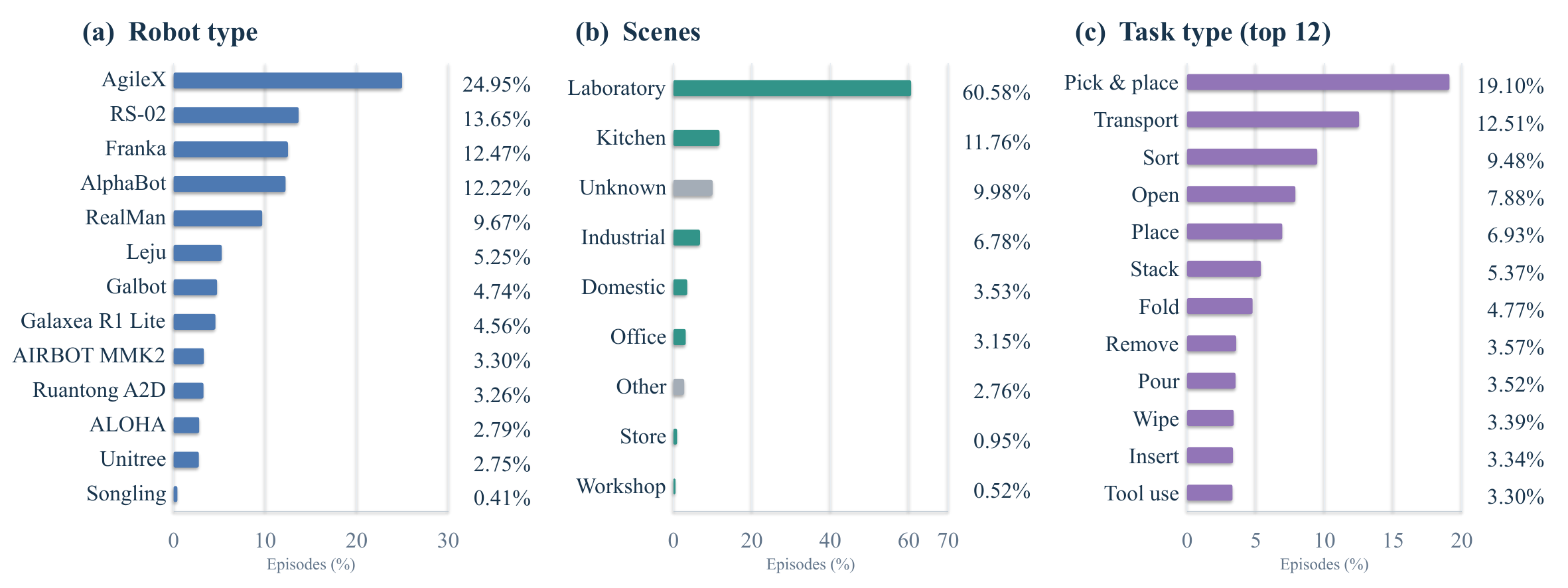}
  \caption{\textbf{Coverage of the 150,000 curated clips.} The panels show the share of episodes by robot family, scene, and task. The task panel shows the twelve most frequent labels. An episode can contribute to more than one task.}
  \label{fig:supp_distribution}
\end{figure}

\Cref{fig:supp_distribution} shows the resulting coverage. The 500 test episodes are held out by source identity, including all derived views, references, and windows.

\clearpage
\section{Multi-View Model and Training}
\label{supp:video_model}

\subsection{Backbone and Conditioning}
\label{supp:backbone_controls}
We initialize the video VAE, UMT5-XXL text encoder~\citep{chung2023unimax}, and 30-block DiT~\citep{peebles2023dit} from \texttt{alibaba-pai/Wan2.2-Fun-5B-Control}~\citep{alibabapai_wan22fun5bcontrol}.
The primary, left, and right view streams share these components and are batched together, with separate token sequences for each view.
The left and right slots accommodate auxiliary cameras across embodiments, including wrist cameras where available.
The native DiT supplies within-view spatiotemporal attention, 3D rotary position embeddings, and text cross-attention.

\noindent\textbf{Depth and history inputs.}
For each view, the DiT input contains 148 channels: 48 noisy video-latent channels, 48 depth-latent channels, four mask channels, and 48 masked-video latent channels.
We encode depth with the video VAE and concatenate it with the noisy video latents along the channel dimension.
Depth normalization follows \cref{sec:multiview_generation}.
The mask and masked-video channels carry history conditioning; these 52 channels are zero for windows without history.
The DiT predicts a 48-channel video-latent velocity.

\noindent\textbf{Reference and text inputs.}
We encode each reference image with the VAE and project its latent through the native reference convolution into a token prefix.
These tokens interact with video tokens through the DiT's self-attention and are removed before decoding the generated video latents.
UMT5-XXL encodes the shared scene-and-action caption with a maximum length of 512 tokens, and the resulting embeddings condition each stream through text cross-attention.

\subsection{Cross-View Residual Blocks}
\label{supp:cross_view_blocks}
We insert a residual block after each of the 6th, 12th, 18th, and 24th DiT blocks.
Each insertion point has its own parameters, shared across the three views.
We first normalize the 3,072-dimensional video features and project them into a 1,024-dimensional bottleneck.
At a fixed latent time, let $Z_\ell^v=D_\ell H_\ell^v$ denote these projected features and $Z_\ell^{-v}$ the concatenated spatial tokens from the other available views.
The block computes
\begin{equation}
\begin{aligned}
 A_\ell^v &= Z_\ell^v + g_{\ell,a}(t)\odot
 \operatorname{Attn}_\ell(Z_\ell^v,Z_\ell^{-v}),\\
 B_\ell^v &= A_\ell^v + g_{\ell,f}(t)\odot
 \operatorname{FFN}_\ell(\operatorname{LN}(A_\ell^v)),\\
 H_\ell^{v,+} &= H_\ell^v + U_\ell B_\ell^v.
\end{aligned}
\label{eq:supp_cross_view_block}
\end{equation}
Here $D_\ell$ includes the initial normalization and down-projection, and $U_\ell$ projects the bottleneck features back to 3,072 dimensions.
The up-projection receives the complete bottleneck output $B_\ell^v$, including its within-view residual.
We initialize both the weight and bias of $U_\ell$ to zero, following the zero-initialized residual-conditioning principle~\citep{zhang2023controlnet}.

\noindent\textbf{Attention and modulation.}
The target view supplies queries, and all spatial tokens from the other available views supply keys and values at the same latent time.
We use 16 attention heads, per-head query/key RMS normalization with $\epsilon=10^{-6}$, and learned view-role embeddings for the primary, left, and right slots in the queries and keys.
View-role embeddings identify the camera slots; cross-view attention exchanges features without explicit camera calibration or reprojection.
The feed-forward network has hidden width 4,096.
A 128-dimensional sinusoidal encoding of the diffusion timestep $t$ passes through an MLP with widths $128\rightarrow128\rightarrow2048$.
Its output is split into two 1,024-dimensional vectors, transformed by $1+\tanh(\cdot)$ to produce the channel-wise gates $g_{\ell,a}$ and $g_{\ell,f}$.
All views read pre-update features and update synchronously.
Reference-prefix and padding tokens are excluded, and missing view slots are masked.

\begin{table}[!htbp]
  \centering
  \caption{\textbf{Video generator dimensions and parameter counts.} The pretrained DiT remains frozen; joint adaptation trains the four cross-view blocks and shared LoRA adapters.}
  \label{tab:supp_cross_view_config}
  \small
  \begin{tabular}{@{}lr@{}}
    \toprule
    Setting & Value \\
    \midrule
    DiT hidden width & 3,072 \\
    Cross-view bottleneck & 1,024 \\
    Attention heads & 16 \\
    Feed-forward hidden width & 4,096 \\
    Original DiT parameters & 5,001,609,408 \\
    Four cross-view blocks & 76,702,720 \\
    Shared LoRA adapters & 23,592,960 \\
    Trainable parameters in stage 2 & 100,295,680 \\
    \bottomrule
  \end{tabular}
\end{table}

\subsection{Training Details}
\label{supp:training_inputs_objective}
\noindent\textbf{Temporal and spatial sampling.}
We resize complete RGB, depth, and reference frames bilinearly to $640\times352$, without cropping, letterboxing, or flipping.
At each common 24-FPS target timestamp, we select the latest source frame at or before that time.
Training windows contain 121 frames with a stride of 96; incomplete tail windows are dropped.
For each source clip, we sample one synchronized reference timestamp from its common valid time range and reuse the per-view reference images for all windows.
The reference timestamp can lie before, within, or after an individual window.
All views and windows of a clip share its caption.
We shuffle windows during training, so longer clips contribute more windows.

\noindent\textbf{History conditioning.}
In both training stages, approximately 25\% of training windows use their first 25 ground-truth frames as history through the native mask/video pathway.
Only the first latent in the noisy video stream is hard-clamped; the other history positions remain noised and supervised.
The same history condition is applied across the synchronized views.

\noindent\textbf{Noise schedule.}
For the objective in \cref{eq:flow_target,eq:flow_loss}, the timestep index is sampled uniformly from the base 1,000-step schedule with shift 5.

\subsection{Two-Stage Optimization}
\label{supp:staged_optimization}
We train for 20K updates on four nodes with eight A800 GPUs each, using one synchronized multi-view sample per GPU and a global batch of 32.
The original DiT, VAE, and text-encoder weights remain frozen throughout.
In the first 4K updates, we optimize only the cross-view blocks at a constant learning rate of $10^{-4}$; LoRA contributes a zero residual and remains fixed.
In the following 16K updates, we jointly optimize the cross-view blocks and shared rank-32 LoRA adapters~\citep{hu2022lora} in all 30 DiT blocks.
LoRA is restricted to the query, key, value, and output projections of native self-attention.
The cross-view learning rate decays linearly from $10^{-4}$ to $5\times10^{-5}$ over this stage.
The LoRA learning rate increases from zero to $5\times10^{-6}$ over its first 500 updates and then stays constant.
We carry the cross-view optimizer state, sample order, and random-number-generator state across the stage boundary.

\noindent\textbf{Optimizer and numerical settings.}
We use AdamW~\citep{loshchilov2019adamw} with the settings in \cref{tab:supp_training}.
Pretrained weights are stored in BF16, while trainable parameters and optimizer states use FP32; computation uses BF16 autocasting and gradient checkpointing.
Text dropout replaces the caption with an empty prompt for all views in a sample together.
Depth and reference conditioning are retained throughout training.
We evaluate the video generation model directly, without exponential moving averaging.

\begin{table}[!htbp]
  \centering
  \caption{\textbf{Video generator training settings.} Training comprises 4K updates of the cross-view blocks followed by 16K updates jointly optimizing the blocks and LoRA adapters.}
  \label{tab:supp_training}
  \small
  \begin{tabular}{@{}ll@{}}
    \toprule
    Setting & Value \\
    \midrule
    Hardware & $4\times8$ A800 (80 GB) GPUs \\
    Global batch / accumulation & 32 / 1 \\
    Cross-view warm-up & 4K updates \\
    Joint adaptation & 16K updates \\
    LoRA rank / alpha & 32 / 32 \\
    AdamW $(\beta_1,\beta_2)$ & $(0.9,0.999)$ \\
    AdamW $\epsilon$ & $10^{-8}$ \\
    Weight decay & 0.01 \\
    Gradient clipping & 1.0 \\
    Text / LoRA dropout & 0.1 / 0 \\
    \bottomrule
  \end{tabular}
\end{table}

\subsection{Long-Video Inference}
\label{supp:long_video_inference}
We generate each window with 50 FlowMatch Euler steps, classifier-free guidance~\citep{ho2022cfg} scale 6, and schedule shift 5.
All views use the same temporal window and diffusion timestep, with independent noise.
We keep the per-view reference images fixed and advance the depth-control window along the trajectory.
Each continuation window receives the previous window's last 25 generated frames as history, using floating-point RGB outputs before video encoding.
The first latent is hard-clamped and the remaining history provides soft conditioning.
We retain the existing output and discard the new window's first 25 frames, appending its remaining 96 frames.
For the final incomplete window, we repeat the last depth frame to fill the input and trim the generated output to the requested duration.
Output timestamps are retained for pairing with simulator actions and states.

\noindent\textbf{Window indexing.}
For a sequence of $T$ frames, window $k$ starts at frame $96k$ and covers frames $96k$ through $96k+120$. The first window emits 121 frames; each subsequent full window emits 96 new frames after its 25-frame prefix is removed. After $K$ full windows, the assembled sequence has $121+96(K-1)$ frames. Tail padding and output trimming preserve the requested $T$ timestamps.
\subsection{Ablation Study}
\label{supp:ablation_training}
\noindent\textbf{Controlled optimization.}
The four ablation variants are independent single-stage runs initialized from the official base checkpoint.
Each trains for 4,000 updates on 32 A800 GPUs with global batch size 32 and seed 43.
All variants use rank-32 LoRA in native self-attention query, key, value, and output projections; the original DiT, text encoder, and VAE remain frozen.
Where present, the four cross-view blocks are optimized jointly with LoRA from the first update.
Learning rates increase linearly over 500 updates to $5\times10^{-6}$ for LoRA and $10^{-4}$ for the cross-view blocks, then remain constant.
AdamW settings, gradient clipping, text dropout, resolution, and window length follow the formal run.
All comparisons use the final 4,000-update checkpoint.

\noindent\textbf{Matched inputs and variants.}
The ablations use the training data with complete three-view inputs, excluding the held-out evaluation episodes and clips lacking a right view.
Sample order, diffusion timesteps, per-view noise, reference images, text dropout, and LoRA initialization are paired across variants.
The independent-view variant shares one backbone and one set of LoRA weights across three streams, with no cross-view blocks.
The latent-concatenation variant stacks independently encoded video, depth, reference, and history latents along the spatial height dimension in primary/left/right order, rearranging the timestep map accordingly.
The shared DiT processes this combined grid; outputs are split into views for loss computation and independent VAE decoding.
This adapts RoboRender's fusion strategy~\citep{huang2026roborender} to our backbone and controls.
The full residual variant uses the architecture above; its no-history counterpart removes the training prefix condition and uses the ordinary generation loss with the same targets and references.
The other three variants share the same history subset, covering approximately 25\% of sampled windows.
Evaluation protocols are specified in \cref{supp:ablation_protocol}.

\clearpage
\section{Video Evaluation Protocols}
\label{supp:video_protocols}

\begin{table}[H]
\centering
\small
\setlength{\tabcolsep}{4pt}
\caption{\textbf{Evaluation units and outputs.} Each protocol uses its own sampling and aggregation rules.}
\label{tab:supp_evaluation_map}
\begin{tabularx}{\textwidth}{@{}l p{0.25\textwidth} X@{}}
\toprule
Protocol & Evaluation unit & Outputs and aggregation \\
\midrule
Trajectory planning & 768 layouts across nine tasks & 720 successful plans; pooled success over attempted layouts. \\
Paired video & 500 episodes; 1,424 matched views & FID/FVD from pooled features; PSNR/SSIM/LPIPS averaged through frames, views, and episodes. \\
Video ablations & 500 episodes; 1,500 views & Full-frame reconstruction, final-segment LPIPS, and overlap L1; equal episode/view weighting. \\
Human preference & 50 task--scene pairs; 12 experts & Five complete rankings per question; mean rank, Top-1/2, and evaluator-bootstrap intervals. \\
Physical default scene & Four tasks; three data sources & Ten trials per task/source; task-completion rate and equal-task average. \\
Physical generalization & Two tasks; five conditions & Ten trials per task/condition/source; condition and task averages. \\
\bottomrule
\end{tabularx}
\end{table}

\subsection{Paired Real-Video Evaluation}
\label{supp:paired_protocol}
The paired set contains 500 held-out episodes and 1,424 camera views matched across all four systems.
Every episode contributes a primary view; 462 also contribute two wrist views, for a total of 924 wrist views.
The remaining 38 episodes contribute only their primary view to the matched evaluation.
We report results for the primary, wrist, and all-view subsets in \cref{tab:real_eva_comparison}.

\noindent\textbf{Alignment and sampling.}
We evaluate all methods at a common $416\times240$ spatial crop with aligned temporal samples.
We use RoboRender's centered crop in normalized ground-truth coordinates, mapping full-frame outputs to the corresponding crop and resizing them to the same extent.
We exclude the initial frame.
For image metrics, we uniformly select 32 frames per view from the intersection of valid source-frame indices shared by the generated outputs and reference recording.
FVD uses a centered 64-frame clip at 8 FPS, aligned through the nearest common presentation timestamps and repeating frames when necessary.
PSNR, SSIM~\citep{wang2004ssim}, and LPIPS~\citep{zhang2018lpips} are aggregated from frames to views and then to episodes.
FID~\citep{heusel2017fid} and FVD~\citep{unterthiner2018fvd} are computed from pooled feature distributions for each subset.
We compute all-view scores directly using these aggregation rules, rather than averaging subset scores by their view counts.

\noindent\textbf{Metric implementations.}
FID uses the 2,048-dimensional Inception pool3 features from \texttt{pytorch-fid} 0.3.0.
FVD uses the StyleGAN-V implementation~\citep{skorokhodov2022styleganv} with its I3D~\citep{carreira2017i3d} TorchScript extractor and 400-dimensional pre-softmax features.
LPIPS uses version 0.1.4 with AlexNet and the v0.1 weights.
SSIM is averaged over RGB channels with an 11-pixel Gaussian window, standard deviation 1.5, population covariance, and data range 255.

\noindent\textbf{Conditioning and comparison scope.}
We compare the four systems using the following conditioning and generation configurations.
RoboRender~\citep{huang2026roborender} uses depth, captions, and source RGB with the robot foreground retained and the background masked by SAM~3~\citep{carion2025sam3}; it generates the available views jointly and does not use the real first frame as a target-image condition.
It uses 81-frame windows with one-frame overlap at $416\times240$ and 15 FPS, with 50 denoising steps, CFG 5, and seed 42.
Cosmos-Transfer2.5~\citep{nvidia2025cosmostransfer25} uses its General Depth configuration independently for each view, with depth and captions but no source RGB, target first frame, or robot mask.
It runs at the 480p setting with 20 denoising steps and uses the previous generated endpoint to continue later windows.
The VDA-S relative-depth controls in this paired set use full-clip normalization.
The Cosmos pipeline includes uint8/H.264 encoding and integer rescaling of depth; we observed ripple artifacts in inspected examples.
Wan-Fun-Control~\citep{alibabapai_wan22fun5bcontrol} uses the original 5B weights without LoRA or cross-view modules, generating each view independently in one full-sequence pass at 352p.
Its controls are the caption, the corresponding depth video, and the real first-frame reference; sampling uses 50 steps, CFG 6, and seed 43.
Our video generation model generates synchronized views at $640\times352$ and 24 FPS, using 121-frame windows with 25-frame history overlap, 50 steps, CFG 6, and schedule shift 5.
The real first-frame references remain fixed across windows, while the depth and generated history advance along the trajectory.
We use seed $43+k$ for the zero-based window index $k$.
The four-system comparison evaluates the complete generation pipelines; the matched ablations below test architectural and history-conditioning choices.

\subsection{Open-Scene RoboTwin Evaluation}
\label{supp:open_protocol}

We use 50 tasks from RoboTwin 2.0~\citep{chen2026robotwin2}, with one trajectory and three synchronized camera views per task.
The benchmark spans ALOHA-AgileX, PiPER, Franka Panda, ARX-X5, and UR5-WSG configurations.
Fifteen scene references are grouped into laboratory, home, and natural-outdoor categories, with five references per category.
Each trajectory receives one scene assignment, yielding 17 laboratory, 17 home, and 16 outdoor questions.
The evaluation unit is one assigned task--scene pair.
Target appearance images are generated references, not ground-truth real observations of the simulated tasks.
All methods use the benchmark-specified per-view target references through their respective conditioning interfaces.
\Cref{fig:supp_benchmark} summarizes the evaluation.

\noindent\textbf{Generation configurations.}
RoboRender jointly generates the three views from simulated depth, SAM~3-extracted robot RGB, captions, and the per-view target images for the first window.
It uses 81-frame windows with one-frame overlap, $416\times240$ resolution, 15 FPS, 50 denoising steps, and CFG 5; later windows are continued from generated endpoints.
Cosmos-Transfer2.5 generates each view independently at 480p with 20 steps and CFG 3.
The target image is supplied through \texttt{image\_context} in each window, without forcing it to be the first output frame; captions and simulated depth provide the other controls.
Both baseline runs use seed 42.
Their simulation-depth inputs linearly map 0.1--3\,m for the primary camera and 0.02--1\,m for the wrist cameras, with clipping outside these ranges.
Our depth audit identified clipping in 22 of the 100 wrist streams, affecting 16 questions.
The reported 50-question results use these inputs; small inverse-depth diagnostic runs were not included.
Wan-Fun-Control uses the original weights and independent full-sequence generation with captions, depth, and per-view scene references, at 15 FPS, $640\times352$ for the primary view and $512\times384$ for wrist views.
It uses 50 steps, CFG 6, and seed 43.
Its depth inputs use the same fixed camera-specific ranges described above.
We evaluate our video generation model with the multi-view windowed sampling configuration used in the paired benchmark.
We keep the per-view target appearance images fixed as references throughout the sequence.
For this simulation benchmark, we convert metric depth to inverse depth and apply min--max normalization independently over each view's full sequence, following the relative-depth encoding used in training.
We mask invalid depths and encode them as black.
The comparison uses each pipeline's stated depth preprocessing and reference interface.

\noindent\textbf{Human protocol.}
Twelve expert evaluators each complete the same 50 questions.
We present the four systems anonymously and independently randomize their positions for each evaluator and question, while keeping the question order fixed.
For every question, the evaluator supplies a complete ranking from first to fourth on each of five dimensions: multi-view consistency, physical plausibility, visual quality, temporal coherence, and task-and-scene adherence.
This yields 3,000 complete rankings, with 600 rankings per dimension.
Mean rank weights all five dimensions equally; Top-1 and Top-2 measure the proportions of rankings in which a method is first or among the first two.
In the plots, ``scene adherence'' abbreviates the task-and-scene criterion.

\noindent\textbf{Uncertainty.}
We compute 95\% percentile confidence intervals for first-place shares and mean ranks using 10,000 evaluator-cluster bootstrap replicates~\citep{efron1979bootstrap} (seed 20261022).
In each replicate, we draw 12 evaluators with replacement and retain their complete rankings of all four methods across the 50 questions and five criteria.
We recompute each statistic, averaging criteria equally for the overall first-place share, and use the 2.5th and 97.5th percentiles as interval endpoints.
The bootstrap unit is the evaluator, with the 50 task--scene pairs held fixed.

\subsection{Video Ablation Evaluation}
\label{supp:ablation_protocol}

\noindent\textbf{Setup.}
We evaluate all three views of the same 500 held-out source episodes, yielding 1,500 view streams compared with the 1,424 matched views in the main comparison.
The sequences last 10--19.67\,s, yielding 1,779 synchronized windows per variant; all episodes require continuation.
Every variant uses the same depth controls, captions, fixed first-frame references, and per-view sampling noise.
Generation uses $640\times352$ frames at 24 FPS, 121-frame windows with 25-frame overlap, 50 denoising steps, CFG 6, schedule shift 5, and seed $43+k$ for window $k$.
Noise is sampled per view before packing the latent-concatenation inputs.
All four variants, including the one trained without history, condition continuation on their own generated tails and discard the regenerated 25-frame prefix when assembling the output.

\noindent\textbf{Evaluation metrics.}
We compare full output frames with the corresponding resized real observations at $640\times352$.
We compute PSNR and SSIM for every output frame, including the initial frame.
LPIPS uses AlexNet v0.1, sampled every six frames (4 FPS), with additional samples at continuation boundaries and the last frame.
For each metric, we average frame scores within each episode and view, then average equally over episodes and the three views.
Final-segment LPIPS restricts the sampled frames to the newly emitted portion of the last window, excluding the duplicated prefix and any padded frames beyond the sequence end.
The ablations use full-resolution, three-view evaluation; the paired benchmark uses the matched-view crop protocol above.

\noindent\textbf{Overlap agreement and scope.}
Overlap L1 is the mean absolute RGB difference between a continuation window's regenerated prefix and the preceding window's generated tail, using original float32 values in $[0,1]$.
It averages the 24 overlapping frames after the first hard-constrained frame, then averages across continuation windows within each episode, episodes, and views.
The table reports this error in units of $10^{-2}$.
Overlap L1 evaluates temporal agreement in the regenerated prefix before it is discarded during assembly.

\subsection{Aggregation Units}
For a framewise metric $m$, let $\mathcal{T}_{ev}$ be the evaluated frames of view $v$ in episode $e$. We first compute
\begin{equation}
 \bar m_{ev}=\frac{1}{|\mathcal{T}_{ev}|}\sum_{t\in\mathcal{T}_{ev}}m_{evt},\qquad
 \bar m=\frac{1}{|\mathcal E|}\sum_{e\in\mathcal E}\frac{1}{|\mathcal V_e|}\sum_{v\in\mathcal V_e}\bar m_{ev}.
\end{equation}
Here $\mathcal V_e$ is the available view subset for that evaluation. Each episode receives equal weight. FID and FVD instead fit feature distributions to pooled image and video samples, respectively.

For human evaluation, let $r_{eqc}^{(m)}\in\{1,2,3,4\}$ be method $m$'s rank from evaluator $e$ on question $q$ and criterion $c$. With $N=12\times50\times5=3{,}000$, the overall summaries are
\begin{equation}
 R^{(m)}=\frac{1}{N}\sum_{e,q,c}r_{eqc}^{(m)},\qquad
 P_k^{(m)}=\frac{1}{N}\sum_{e,q,c}\mathbf1[r_{eqc}^{(m)}\leq k].
\end{equation}
Top-1 and Top-2 report $100P_1$ and $100P_2$. Per-criterion summaries use the corresponding 600 rankings; pairwise preference is the share of rankings in which the row method precedes the column method.

\clearpage
\section{Additional Video Results}
\label{supp:video_results}

\begin{figure}[H]
  \centering
  \includegraphics[width=1.0\textwidth]{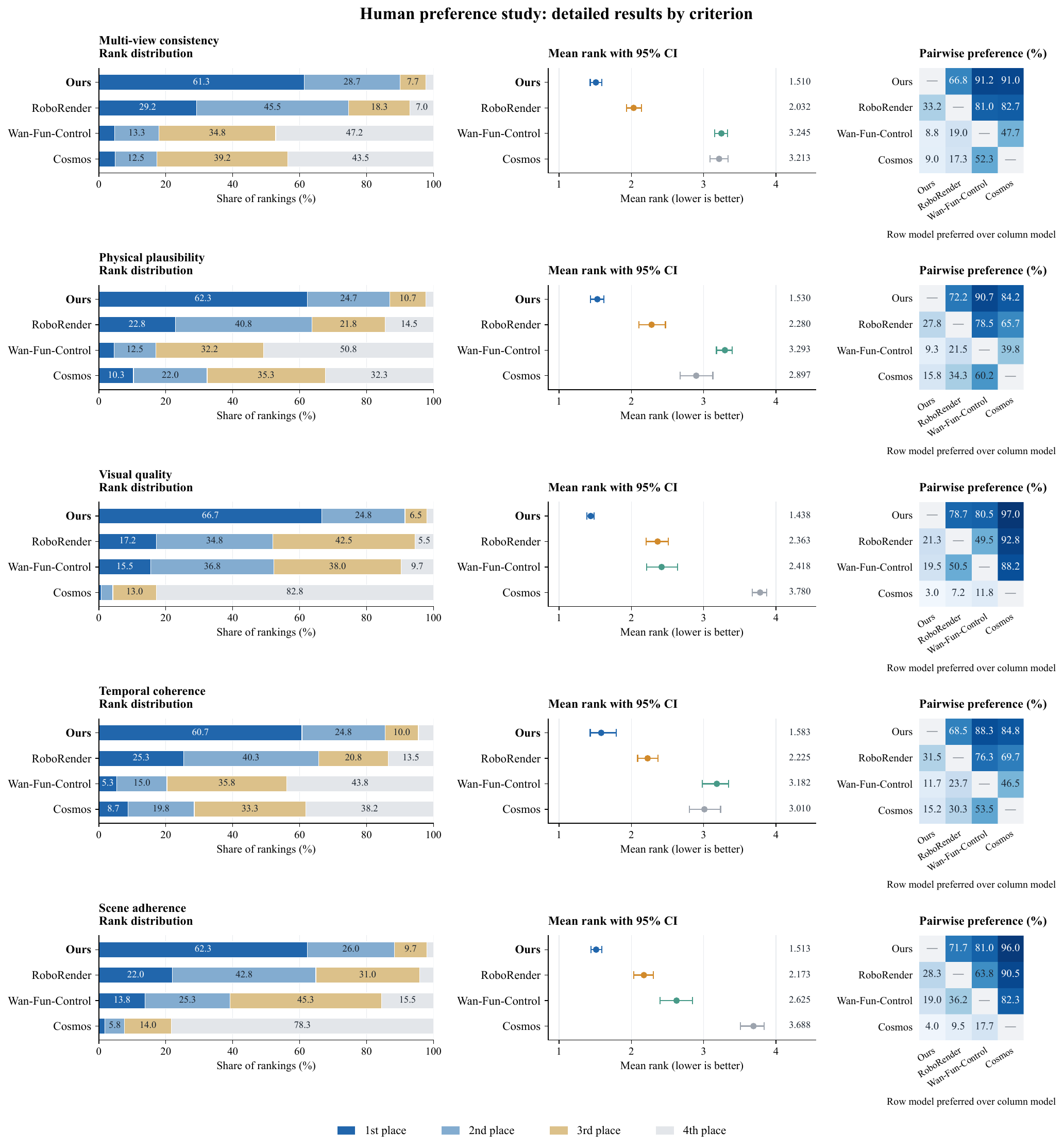}
  \caption{\textbf{Human preference by criterion.} Left: rank distributions. Middle: mean ranks with 95\% evaluator-cluster bootstrap confidence intervals (lower is better), using the protocol in \cref{supp:open_protocol}. Right: pairwise preference rates for the row method over the column method.}
  \label{fig:supp_preference_detail}
\end{figure}

\noindent\textbf{Preference by criterion.}
Our model's mean ranks range from 1.438 to 1.583. \Cref{fig:supp_preference_detail} shows rank distributions, evaluator-bootstrap intervals, and pairwise preferences. ``Cosmos'' denotes Cosmos-Transfer2.5.

\clearpage
\section{Additional Qualitative Comparisons}
\label{supp:video_cases}
One held-out real-video example covers ALOHA; two further examples vary simulated scenes.
Each panel compares depth inputs and four systems across three views and timestamps.

\par\medskip
\noindent\begin{minipage}{\linewidth}
  \centering
  \includegraphics[width=\linewidth]{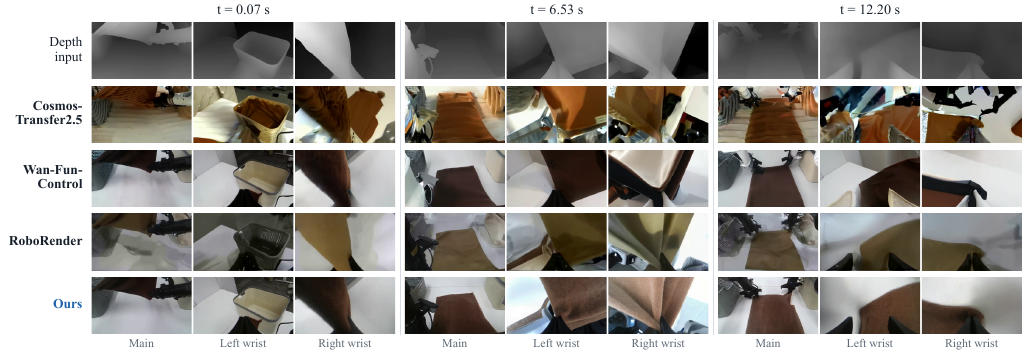}
  \captionof{figure}{\textbf{Held-out real-video example: ALOHA.} Cloth and storage bin.}
  \label{fig:supp_case_aloha}
\end{minipage}\par
% \par\vfill
% \noindent\begin{minipage}{\linewidth}
%   \centering
%   \includegraphics[width=\linewidth]{figure/cases/eva_500_case/case_C08_ruantong_a2d.pdf}
%   \captionof{figure}{\textbf{Held-out real-video example: Ruantong A2D.} Cluttered tabletop workspace.}
%   \label{fig:supp_case_ruantong}
% \end{minipage}\par
% \par\vfill
% \noindent\begin{minipage}{\linewidth}
%   \centering
%   \includegraphics[width=\linewidth]{figure/cases/eva_500_case/case_C10_unitree.pdf}
%   \captionof{figure}{\textbf{Held-out real-video example: Unitree.} Grapes and a bowl on a grid-patterned mat.}
%   \label{fig:supp_case_unitree}
% \end{minipage}\par
% \clearpage
\noindent\begin{minipage}{\linewidth}
  \centering
  \includegraphics[width=\linewidth]{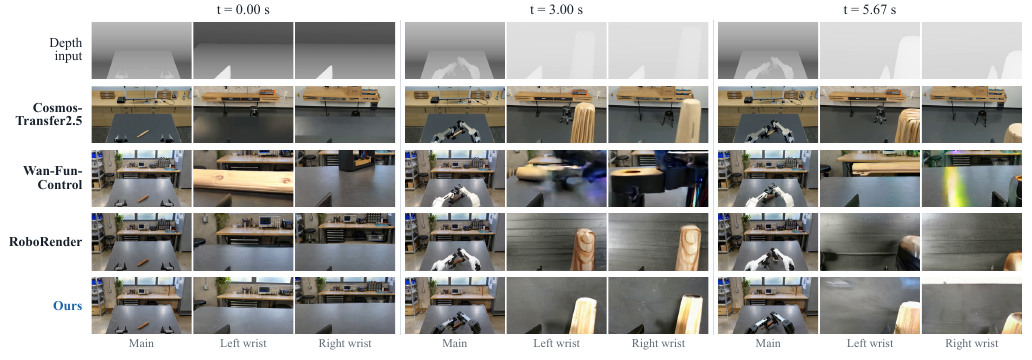}
  \captionof{figure}{\textbf{Open-scene example: workbench.} Wooden object on a workbench.}
  \label{fig:supp_case_q008}
\end{minipage}\par
\par\vfill
\noindent\begin{minipage}{\linewidth}
  \centering
  \includegraphics[width=\linewidth]{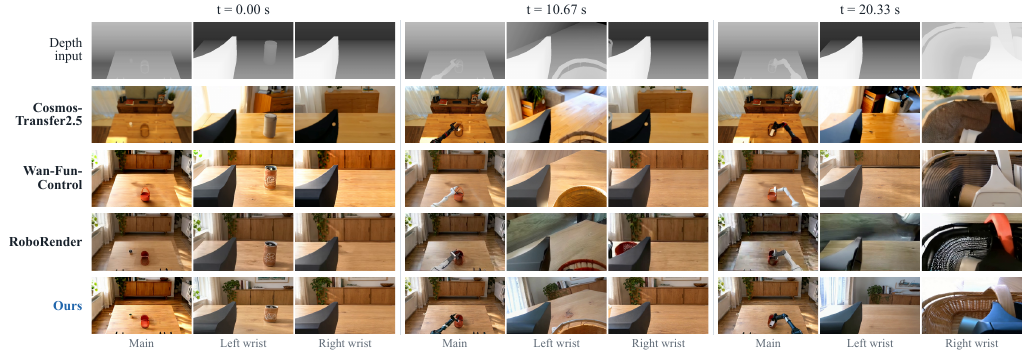}
  \captionof{figure}{\textbf{Open-scene example: home environment.} Tabletop containers under natural lighting.}
  \label{fig:supp_case_q026}
\end{minipage}\par
% \par\vfill
% \noindent\begin{minipage}{\linewidth}
%   \centering
%   \includegraphics[width=\linewidth]{figure/cases/eva_reference_case/case_Q030.pdf}
%   \captionof{figure}{\textbf{Open-scene example: laboratory.} Blue container on a light tabletop.}
%   \label{fig:supp_case_q030}
% \end{minipage}\par
\clearpage

\FloatBarrier
\section{Real-World Experimental Details}
\label{supp:real_world_tasks}
We compare three training data sources on four tasks across Astribot, AgileX, and Tienyi 2.0, then evaluate visual generalization on the two Astribot tasks. Task completion is judged manually from the complete execution.

\subsection{Default-Scene Comparison}
\emph{Reconstructed Sim Data} pairs simulated observations and actions from reconstructed scenes. \emph{\method Synthetic Data} pairs generated multi-view observations with the corresponding simulator actions. \emph{Real Robot Data} consists of teleoperated demonstrations. Each policy uses 80 demonstrations per task in the default scene.

We perform ten rollouts per task and training source, totaling $4\times3\times10=120$ trials. Initial object positions are randomized within the spatial range covered by the training data; task fixtures remain fixed. A rate is $100s/10$, where $s$ counts fully completed trials. The average in \cref{tab:real_world_main} weights the four tasks equally.

\noindent\textbf{Policy training.}
All policies fully fine-tune $\pi_{0.5}$~\citep{physicalintelligence2025pi05} for 30K steps on eight NVIDIA A800 GPUs, with batch size 256 and action chunk length 50.

\subsection{Task Completion Criteria}

Each trial must complete all required stages:

\begin{enumerate}
    \item \textbf{Astribot: Lemon Placement.} The right arm grasps the lemon, transports it, and releases it into the box.
    \item \textbf{Astribot: Cup Handover.} The right arm picks up the cup from the stand and transfers it to the left arm, which places it to the left of the stand. The default cup is blue.
    \item \textbf{AgileX: Drawer Loading.} The left arm opens the drawer, the right arm places a mangosteen inside, and the left arm closes it.
    \item \textbf{Tienyi 2.0: Corn Placement.} The right arm grasps the corn, transports it, and releases it into a fixed basket.
\end{enumerate}

\begin{figure}[!htbp]
  \centering
\includegraphics[width=\linewidth,trim=0 246 0 0,clip]{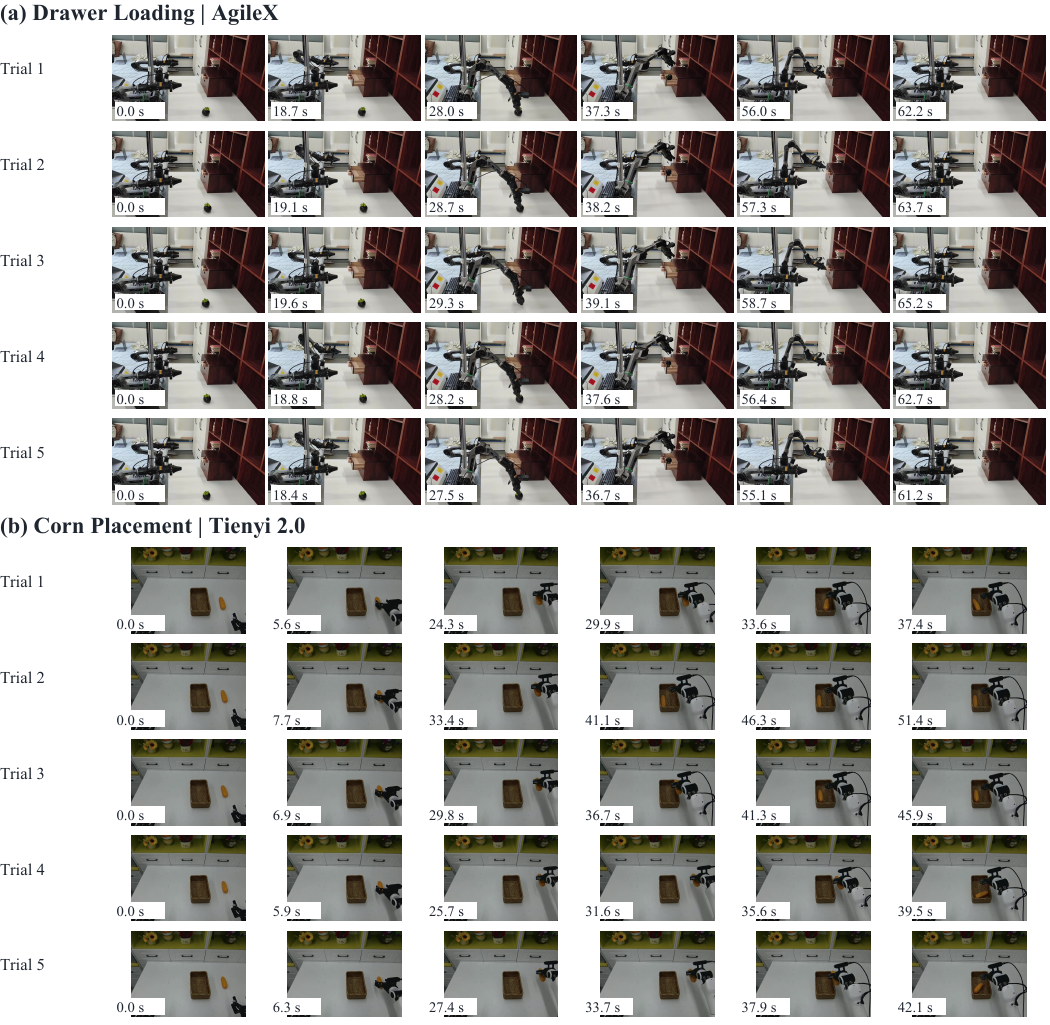}
\caption{\textbf{Successful Drawer Loading on AgileX.} Five trials of the policy trained with \method synthetic data. Each row shows six chronological frames through opening, fruit insertion, and closure; the final column shows the closed drawer. Timestamps are local to each execution.}
\label{fig:supp_real_drawer_corn}
\end{figure}

\begin{figure}[!htbp]
  \centering
\includegraphics[width=\linewidth,trim=0 0 0 246,clip]{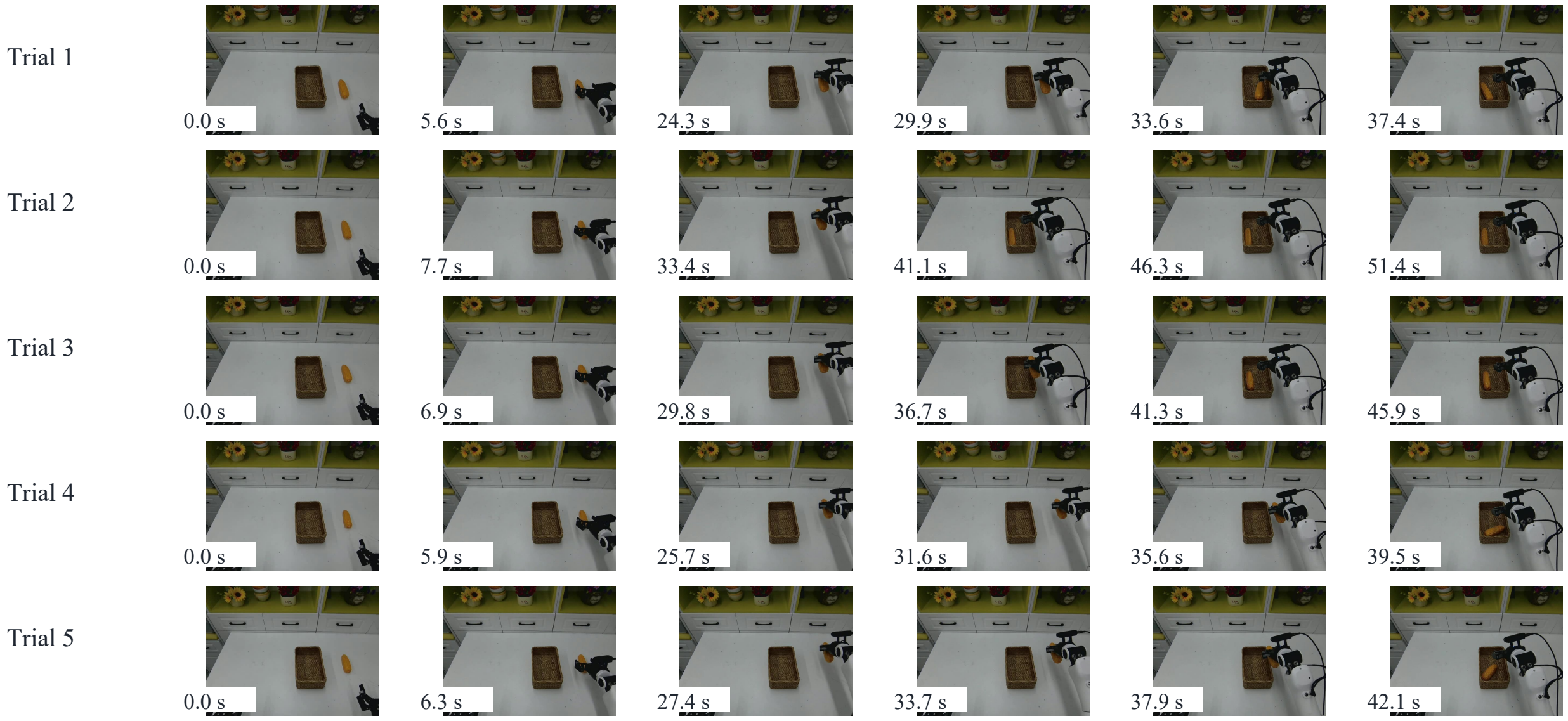}
\caption{\textbf{Successful Corn Placement on Tienyi 2.0.} Five trials of the policy trained with \method synthetic data. Each row shows six chronological frames through grasping, transport, and release; the final column shows the corn inside the basket. Timestamps are local to each execution.}
\label{fig:supp_real_corn}
\end{figure}

\subsection{Visual Generalization on Astribot}
For each Astribot task, the synthetic training set contains 80 demonstrations: 30 in the default scene and 10 in each of five appearance conditions. The comparator uses 80 teleoperated real demonstrations collected in the default scene. Both policies use the same training configuration and task completion criteria.

\noindent\textbf{Light.}
Red light, Green light, and Blue light are evaluated separately for both tasks.

\noindent\textbf{Background.}
A green checkered tablecloth changes the tabletop appearance while retaining the target object and task.

\noindent\textbf{Clutter.}
Lemon Placement adds other fruits as distractors; the robot must still select the lemon. Cup Handover uses yellow, green, or red cups of unchanged shape. Its ten training demonstrations span all three colors, and its ten evaluation rollouts form one condition.

\noindent\textbf{Evaluation.}
Each condition changes one visual factor, with initial object positions randomized within the training-data coverage. The synthetic mixture includes all five appearance conditions used in evaluation. We perform ten rollouts per task, source, and condition, totaling $2\times2\times5\times10=200$ trials. Success is judged manually by the task criteria above, so each condition-specific rate changes in ten-point increments.

\begin{figure}[!htbp]
  \centering
  \includegraphics[width=\linewidth,height=.84\textheight,keepaspectratio]{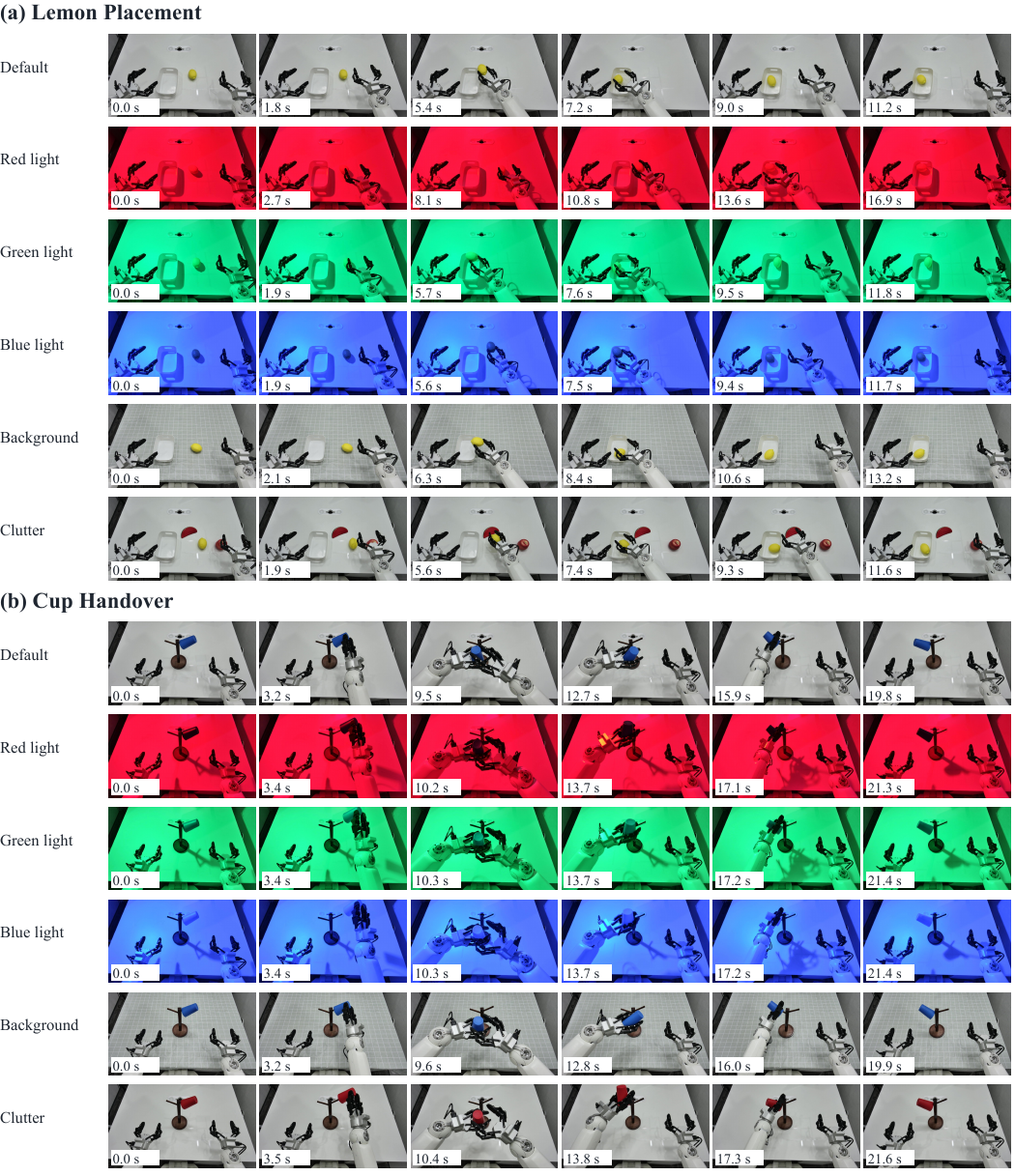}
  \caption{\textbf{Physical rollouts of policies trained with \method data.} Six head-camera frames show selected completed Lemon Placement and Cup Handover rollouts. Condition names match the main figure: Red light, Green light, Blue light, Background, and Clutter. Background changes the tablecloth; Clutter adds fruit distractors or changes the cup color. }
  \label{fig:supp_real_lemon}
  \label{fig:supp_real_cup}
\end{figure}

\noindent\textbf{Aggregation.}
We average the five condition rates within each task, then weight both tasks equally. Lemon Placement averages 50\% for real-data training and 90\% for \method; Cup Handover averages 28\% and 46\%. The overall means are 39\% and 68\%, respectively (\cref{tab:real_world_generalization}). Further completed Lemon Placement and Cup Handover executions appear in \cref{fig:supp_real_lemon}.

\clearpage
\section{Generated Training Data and Physical Rollouts}
\label{supp:dataset_rollout_cases}
Generated training episodes cover four manipulation tasks. Columns follow time; the three camera views share timestamps within each episode.
\par\noindent\begin{minipage}{\linewidth}
  \centering  \includegraphics[width=\linewidth,height=.84\textheight,keepaspectratio]{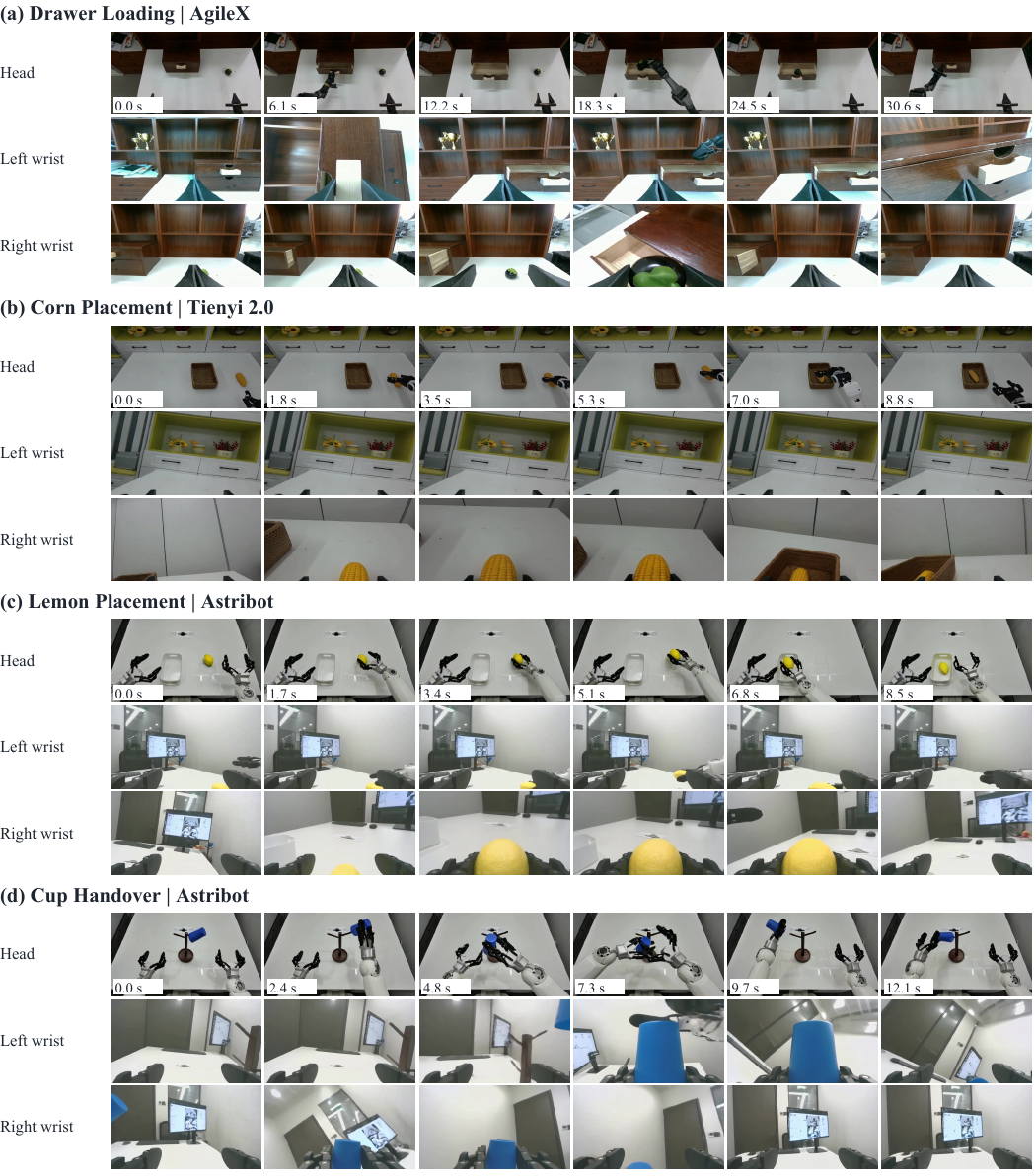}
  \captionof{figure}{\textbf{Generated multi-view training episodes.} Drawer Loading, Corn Placement, Lemon Placement, and Cup Handover, each shown at six timestamps from the head and both wrist cameras. These panels show generated observations, not physical policy rollouts.}
  \label{fig:supp_dataset_tasks_1}
  \label{fig:supp_dataset_tasks_2}
\end{minipage}\par

% \begin{figure}[!htbp]
%   \centering
%   \includegraphics[width=\textwidth]{figure/appendix-showcase/dataset_tasks_extra.pdf}
%   \caption{\textbf{Additional generated episodes across all four tasks.} A second episode for each task, shown at six synchronized timestamps from the head, left-wrist, and right-wrist cameras. Together with the preceding examples, these panels show variation across generated episodes.}
%   \label{fig:supp_dataset_extra}
% \end{figure}
\begin{figure}[!htbp]
  \centering
  \includegraphics[width=\textwidth]{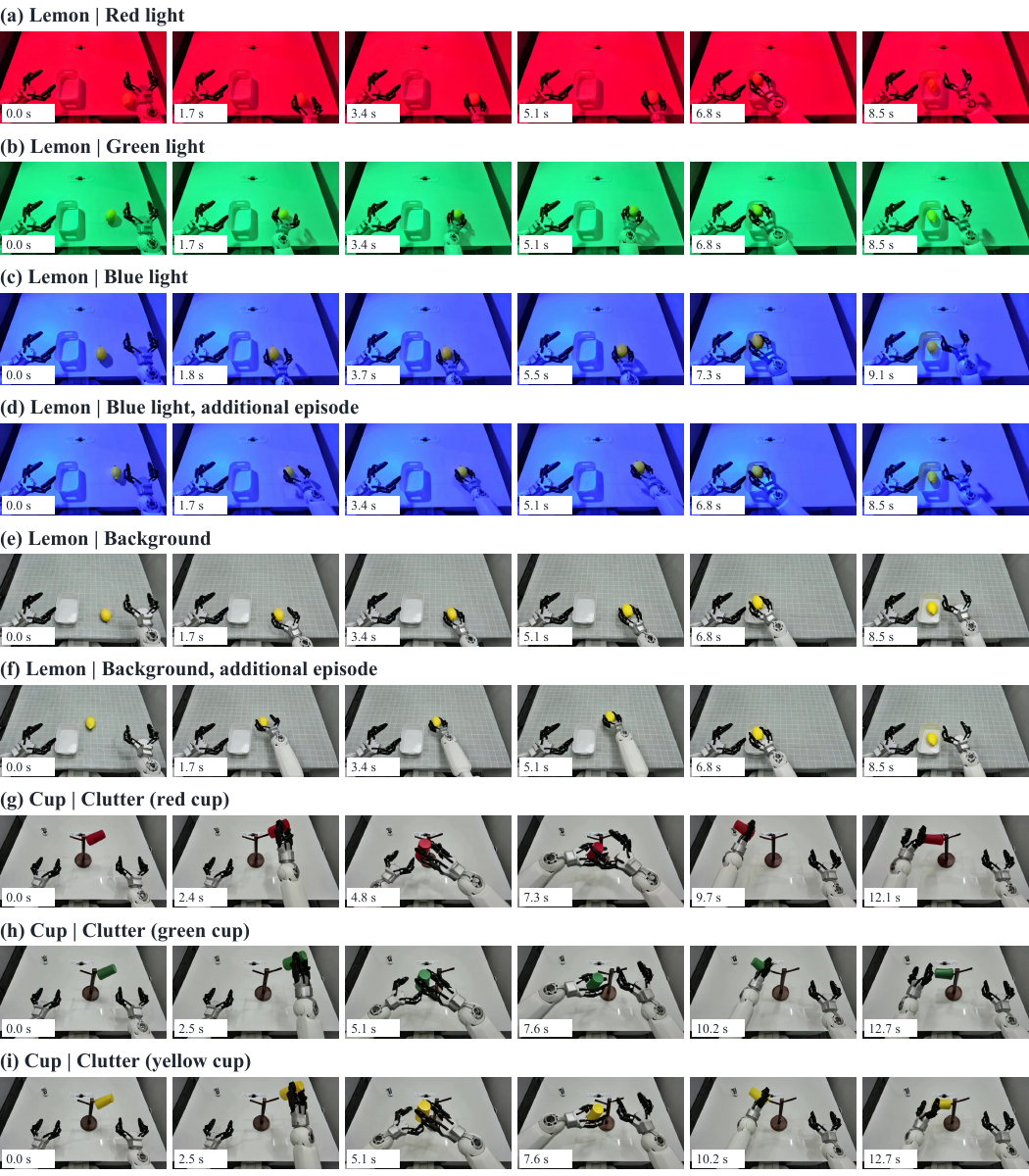}
  \caption{\textbf{Appearance variation in generated training videos.} Head-camera sequences show Light (colored illumination), Background (a checkered tabletop), and Clutter (cup-color variation), following the condition names in the main figure. Each row is a separate episode sampled at six times; rows are not frame-aligned versions of a single trajectory.}
  \label{fig:supp_dataset_appearance}
\end{figure}

% \begin{figure}[!htbp]
%   \centering
%   \includegraphics[width=\textwidth]{figure/appendix-showcase/dataset_generalization_multiview.pdf}
%   \caption{\textbf{Generated multi-view training data for visual generalization.} Additional lemon episodes cover Red light, Green light, and Background; a red-cup episode illustrates Clutter. Each example shows six synchronized timestamps from the head and both wrist cameras. These generated episodes illustrate the appearance coverage provided to policy training; they are distinct from the physical evaluation rollouts below.}
%   \label{fig:supp_generated_generalization_multiview}
% \end{figure}
\FloatBarrier
\subsection{Physical Policy Rollouts}
We show selected completed trials. Each row follows one execution; timestamps are local to the clip, and the last column shows its terminal state. Aggregate results are reported in \cref{supp:real_world_tasks}.

% \begin{figure}[!htbp]
%   \centering
%   \includegraphics[width=\textwidth]{figure/appendix-showcase/real_training_sources.pdf}
%   \caption{\textbf{Additional physical rollouts across training data sources.} Real, Sim, and RoboExpand denote policies trained with real robot demonstrations, reconstructed simulation data, and \method synthetic data, respectively. Two completed trials per source are shown for each task.}
%   \label{fig:supp_real_training_sources}
% \end{figure}

\end{document}